\PassOptionsToPackage{round}{natbib}

\documentclass{article}
\usepackage[preprint]{neurips_2026}

\usepackage{microtype}
\usepackage{graphicx}
\usepackage{subcaption}
\usepackage{booktabs}
\usepackage[dvipsnames]{xcolor}
\usepackage{caption}
\usepackage{makecell}
\usepackage{wrapfig}
\usepackage{graphics}
\usepackage{enumitem}
\usepackage{amsmath}
\usepackage{amssymb}
\usepackage{mathtools}
\usepackage{amsthm}
\let\svthefootnote\thefootnote
\newcommand\freefootnote[1]{%
  \let\thefootnote\relax%
  \footnotetext{#1}%
  \let\thefootnote\svthefootnote%
}
\usepackage{multirow}
\usepackage{algorithm}
\usepackage{algpseudocode}
\usepackage{duckuments}
\usepackage{pifont}
\usepackage{tcolorbox}
\tcbuselibrary{skins, breakable, theorems}
\usepackage[titletoc]{appendix}
\usepackage{listings}
\newtcolorbox{promptbox}[1][]{
  colback=gray!5,
  colframe=gray!50,
  coltitle=black,
  colbacktitle=yellow!20,
  fonttitle=\bfseries,
  title=Prompt,
  enhanced,
  sharp corners=south,
  boxrule=0.8pt,
  left=6pt,
  right=6pt,
  top=6pt,
  bottom=6pt,
  width=\linewidth,
  fontupper=\ttfamily\small,
  before skip=6pt,
  after skip=6pt,
  #1
}

\newtcolorbox{llmbox}[1][]{
  colback=gray!5,
  colframe=gray!50,
  coltitle=black,
  colbacktitle=blue!15,     
  fonttitle=\bfseries,
  title=LLM Output,
  enhanced,
  sharp corners=south,
  boxrule=0.8pt,
  left=6pt,
  right=6pt,
  top=6pt,
  bottom=6pt,
  width=\linewidth,
  fontupper=\ttfamily\small,
  before skip=6pt,
  after skip=6pt,
  #1
}
\newtcolorbox{llmbox2}[1][]{
  colback=gray!5,
  colframe=gray!50,
  coltitle=black,
  colbacktitle=green!15,
  fonttitle=\bfseries,
  title=LLM Output (Alt),
  enhanced,
  sharp corners=south,
  boxrule=0.8pt,
  left=6pt,
  right=6pt,
  top=6pt,
  bottom=6pt,
  width=\linewidth,
  fontupper=\ttfamily\small,
  before skip=6pt,
  after skip=6pt,
  #1
}
\newtcolorbox{outputbopx}[1][]{
  colback=gray!5,
  colframe=gray!50,
  coltitle=black,
  colbacktitle=red!15,
  fonttitle=\bfseries,
  title=LLM Output (Alt),
  enhanced,
  sharp corners=south,
  boxrule=0.8pt,
  left=6pt,
  right=6pt,
  top=6pt,
  bottom=6pt,
  width=\linewidth,
  fontupper=\ttfamily\small,
  before skip=6pt,
  after skip=6pt,
  #1
}

\usepackage{xspace} 

\makeatletter
\let\blx@rerun@biber\relax
\makeatother

\newcommand{\N}{\ensuremath{N}\xspace}
\newcommand{\D}{\ensuremath{D}\xspace}
\newcommand{\Dp}{\ensuremath{D'}\xspace}
\newcommand{\TTP}{\ensuremath{\mathrm{TPP}}\xspace}
\newcommand{\Loss}{\ensuremath{\mathcal{L}}\xspace}
\newcommand{\A}{A\xspace}
\newcommand{\B}{\ensuremath{B}\xspace}
\newcommand{\E}{\ensuremath{E}\xspace}
\newcommand{\C}{\ensuremath{C}\xspace}
\newcommand{\K}{\ensuremath{K}\xspace}

\makeatletter
\let\blx@rerun@biber\relax
\makeatother

\newtcolorbox{definitionbox}[1][]{
  enhanced,
  colback=gray!10,
  colframe=gray!80,
  coltitle=black,
  fonttitle=\bfseries,
  title=Definition,
  #1
}

\usepackage{graphicx}
\usepackage{subcaption}

\usepackage{amsfonts}
\usepackage{nicefrac}

\usepackage{booktabs}

\usepackage{microtype}

\usepackage{url}
\usepackage{hyperref}
\usepackage{cleveref}

\crefname{figure}{Figure}{Figures}        
\Crefname{figure}{Figure}{Figures}        
\crefname{section}{Section}{Sections}
\Crefname{section}{Section}{Sections}
\crefname{equation}{Eq.}{Eqs.}

\hypersetup{
    colorlinks=true,
    urlcolor=blue,
    linkcolor=blue,
    citecolor=blue
}

\newcommand{\strat}{\ensuremath{\mathrm{strat}}}

\usepackage{lineno}

\definecolor{darkblue}{rgb}{0, 0, 0.5}

\title{Bridging Compute- and Data-Optimal Pretraining}

\author{%
  Tian Qin\thanks{Equal contributions. Correspondence to \texttt{tqin@g.harvard.edu}, \texttt{hamidieh@mit.edu }} \\
  Harvard University \\
  \And
  Kimia Hamidieh\footnotemark[1] \\
  MIT CSAIL \\
   \And
  David Alvarez-Melis \\
  Harvard University
}

\begin{document}

\maketitle

\begin{abstract}
Classical compute-optimal scaling laws assume an unbounded supply of fresh pretraining data, yet pretraining is increasingly entering the regime where compute is growing faster than high-quality data. We propose \textit{Compute-Data (CD) scaling laws}, a unified framework that bridges \emph{compute-optimal} scaling, in which data scales freely with compute, and \emph{data-optimal} scaling, in which the corpus 
is fixed and compute can grow unbounded. 
CD scaling extends classic scaling by introducing a \emph{token effectiveness function} $\eta$ that quantifies how much a \emph{derived} token, produced for instance by multi-epoch repetition or paraphrasing, is worth relative to a fresh one, ranging from a perfect substitute to no value at all. 
Fitting $\eta$ for two data expansion strategies (multi-epoch repetition, paraphrasing) across model sizes ranging from 14M to 600M parameters on the Dolma-3 corpus, we find that it is far from constant: it depends jointly on model size, the tokens-per-parameter ratio, and the amount of derived data, and saturates as the corpus is expanded. The functional form of the token effectivness function $\eta$ implies that substituting compute for data \emph{diminishes} with both model size and data availability, and partitions training into three operational regimes: compute-bound, data-bound, and 
model-bound, showing that classic compute-optimal allocation is suboptimal across most of the practically relevant regime.\looseness=-1

\end{abstract}
\section{Introduction}

\begin{figure}[t]
    \centering
    \includegraphics[width=1.0\linewidth]{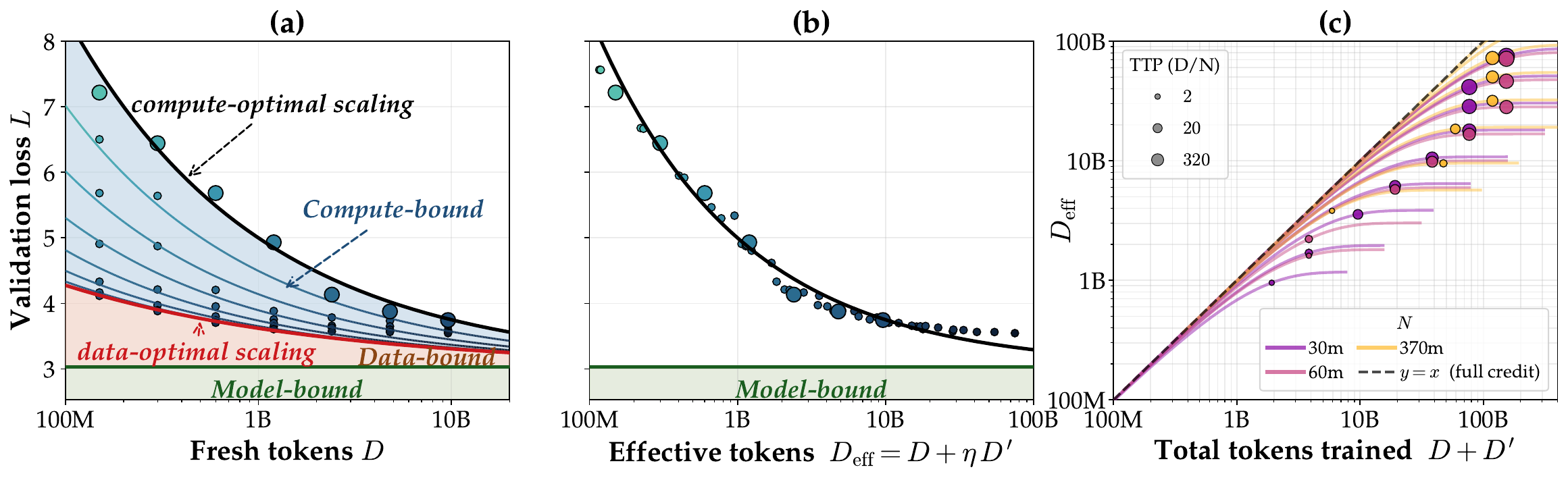}
     \caption{
     \textbf{CD-scaling law overview.}
     \textit{(a)} For a model of size \N, classical scaling laws assume 1-epoch training on fresh data, which coincides with the compute-optimal Pareto frontier. Spending additional compute on \emph{derived} 
tokens \Dp\ produced from \D, via data expansion strategies such as multi-epoch repetition, paraphrasing, or distillation, departs from this frontier and reaches lower loss without curating more fresh data.
     \textit{(b)} To describe training behavior beyond the compute-optimal frontier, CD-scaling laws introduce a token effectiveness function $\eta$ that maps each derived token to a fresh-data equivalent. 
Under this mapping, runs beyond the compute-optimal frontier collapse onto the classical scaling line.
     \textit{(c)} Both the saturation level of effective tokens and the rate of approach depend on \N\ and \TTP: as the model grows and data becomes more abundant, additional compute brings diminishing 
returns.
     }
    \label{fig:main_figure}
    \vspace{-2px}
\end{figure}

The classical scaling laws \citep{hoffmann_training_2022, kaplan_scaling_2020} characterize how pretraining loss decreases with model size \N\ and dataset size \D, prescribing a compute-optimal allocation 
between the two. Implicit in this formulation is a \emph{data-abundance assumption}: fresh tokens are freely available, and the pretraining corpus \D\ scales naturally with compute. But this assumption is becoming inaccurate. Compute continues to grow at exponential rates~\citep{sevilla2022compute}, while high-quality pretraining data remains finite and costly to 
curate~\citep{villalobos2024will}. Pretraining is therefore entering a regime that classical scaling laws were not designed for: one in which \D\ is bounded and additional compute must be spent on 
\emph{derived tokens} \Dp\ produced from \D\ via multi-epoch repetition, paraphrasing, or distillation. The central question becomes:

\vspace{4pt}
  \centerline{\emph{For a model of size \N\ trained on a fixed corpus \D, how much can additional compute reduce loss?}}

We approach this question through the lens of two limiting regimes. \textbf{Compute-optimal scaling} $\Loss^{\mathrm{Chin}}$ describes the setting where data is unbounded and compute is the bottleneck, and it is the regime characterized by classical scaling laws. \textbf{Data-optimal scaling} $\Loss^D$ describes the opposite limit, where \D\ is fixed and compute is unbounded and it is the asymptote of what additional training, in any form, can achieve on a given corpus \D. Modern pretraining sits between these two limits (\Cref{fig:main_figure}\textit{a}).\looseness=-1

We propose \textbf{Compute-Data (CD) scaling laws}, a unified framework that 
bridges the compute- and data-optimal scaling. At its center is an \emph{effectiveness function} $\eta^{\strat}\in [0, 1]$ that quantifies, for a given data expansion strategy such as paraphrasing or repetition, how much a derived token \Dp\ is worth relative to a fresh one (\Cref{fig:main_figure}\textit{b}). 
The resulting scaling law is
\[
\Loss(\N, \D, \Dp) = \E + \frac{\A}{\N^\alpha} + \frac{\B}{(\D + \eta \cdot \Dp)^\beta},
\]

with $\eta = 1$ meaning derived tokens are as informative as fresh data and 
$\eta = 0$ meaning they provide no benefit. The law continuously interpolates 
between the two regimes: when $\Dp = 0$ it recovers compute-optimal scaling, 
and as $\Dp \to \infty$ it approaches the data-optimal limit, predicting the best 
achievable loss for a given $(\N, \D)$.

 We focus on two data-expansion strategies, multi-epoch repetition and paraphrasing. To fit CD-scaling laws, we sweep model sizes \N\ from 14M to 600M parameters, fresh-data sizes \D\ from 30M to 30B tokens, and 
derived-token budgets \Dp\ from 30M to 120B tokens, with hyperparameter search over learning rate and weight decay. We find that the effective-token count $D_{\mathrm{eff}} = \D + \eta\,\Dp$ saturates as $\Dp 
\to \infty$, with both the \textit{saturation level} and \textit{the rate of saturation} depending on \N\ and the tokens-per-parameter ratio $\TTP = \D/\N$ (\Cref{fig:main_figure}\textit{c}). This observation allows us to determine the functional form of the effectiveness coefficient $\eta$, which has a power-law decay over \N, \D\ and \Dp. The data-constrained scaling law of \citet{muennighoff_scaling_2025} is recovered as the special case in which the saturation level is constant in $(\N, \D)$.

The fitted $\eta^{\strat}$ yields three practical implications. First, between the compute-optimal scaling $\Loss^{\mathrm{Chin}}$ and data-optimal scaling $\Loss^D$, the proposed CD scaling law identifies three regimes: \textbf{compute-bound}, \textbf{data-bound}, and \textbf{model-bound}. This informs a practitioner which resource is currently limiting their run. Second, for a given $(\N, \D)$, it prescribes the optimal compute to spend on $\Dp$, and generalizes Chinchilla's compute-only optimization into a \textit{joint} compute--data Pareto frontier. Third, CD-scaling prescribes \emph{which} expansion strategy to use and \emph{how much} additional data to train on. Concretely, extrapolated CD-scaling law shows that the 4-epoch rule holds only for medium-scale models ($\sim 3$B) near $1\times$ Chinchilla, with recommended epochs decreasing as \N\ and \TTP\ grow. Paraphrasing is preferred for small models ($\lesssim 600$M) and sub-Chinchilla data budgets, but is ineffective for $\N \geq 7$B or large budgets ($\geq 4\times$ Chinchilla).\looseness=-1

 In summary, our contributions are:
 \begin{itemize}[itemsep=2pt,labelindent=2pt,topsep=0pt,parsep=0pt,partopsep=1pt,align=left,leftmargin=*]
     \item \textbf{A unified CD-scaling framework.} We propose CD-scaling laws, which bridge the compute-optimal and data-optimal regimes through an effectiveness function $\eta^{\strat}$.
 
     \item \textbf{Effectiveness of derived tokens diminishes with scale.} Fitting $\eta$ on Dolma-3 across model sizes from 14M to 600M, we find that the effectiveness of spending additional compute beyond 
1-epoch training decays with both \N\ and \TTP.

    \item \textbf{Three practical implications.} (a) The CD law identifies three regimes---compute-, data-, and model-bound---that inform which resource is currently limiting the pretraining run. (b) For a given $(\N, \D)$, it prescribes the optimal compute--data allocation, generalizing Chinchilla's compute-only optimization into a joint compute--data Pareto frontier. (c) It prescribes when to switch from repetition to paraphrasing, and provides recommendation on optimal epochs or paraphrasing based on training configurations.

 \end{itemize}

\section{Related Work}
\paragraph{Scaling laws.}
The classical scaling laws of \citet{kaplan_scaling_2020} and \citet{hoffmann_training_2022} characterize loss as a function of model size \N\ and \emph{fresh} pretraining tokens \D, prescribing a compute-optimal allocation under the assumption that \D\ scales freely with compute. Subsequent work has investigated each side of this assumption. \citet{gadre_language_2024} show that the Chinchilla recipe extrapolates reliably into the substantially over-trained regime ($\TTP \gg 20$). \citet{muennighoff_scaling_2025} extend Chinchilla to the data-constrained regime, fitting a single saturation budget for repeated tokens. 
\citet{yan_larger_2025} provide a complementary theoretical perspective, proving in linear regression that the effective reuse rate of repeated data grows with model size \N, with $\Theta(\log \N)$ scaling under strong convexity and power-law scaling under Zipf-distributed features.
\citet{kim_pre-training_2025} take the opposite limit, asking what loss is achievable under fixed \D\ and unbounded compute. Concurrent to our work, \citet{lovelace2026prescriptivescalinglawsdata} also extend Chinchilla to repeated data, but model repetition through an additive, model-size-dependent overfitting penalty that captures the regime where loss \emph{rises} with further epochs, rather than through an effectiveness function on derived tokens. Our work unifies these threads by recovering 
\citet{hoffmann_training_2022} as the $\Dp = 0$ limit, the law of \citet{muennighoff_scaling_2025} as the constant-saturation special case, and the data-optimal limit of \citet{kim_pre-training_2025} as $\Dp \to 
\infty$.
\paragraph{Synthetic data for pretraining.}
A growing body of work treats synthetic data as a lever for trading compute against fresh-token scarcity. \citet{maini_rephrasing_2024} show that rephrasing web documents into structured styles accelerates pretraining by ${\sim}3\times$, and \citet{datologyai_beyondweb_2025} extend the recipe to trillion-token regimes. \citet{kang_demystifying_2025} conduct a controlled study across synthetic-data types, finding that mixing ${\sim}30\%$ rephrased data yields $5$--$10\times$ training speedups. \citet{askari-hemmat_improving_2025} dynamically target synthetic data at the model's current knowledge gaps; 
\citet{yang_synthetic_2025} learn inter-document relations to generate diverse synthetic continuations. Each of these works characterizes a single strategy in isolation. We instead fit a common functional form for the token-effectiveness function $\eta$ across multi-epoch repetition and paraphrasing, and treat each strategy as a parameterization of the same underlying law.
\paragraph{Distillation and other compute-for-data exchanges.}
Distillation \citep{hinton_distilling_2015} has seen renewed interest in LLM pretraining as a way to extract additional signal from a fixed corpus. \citet{kim_pre-training_2025} show that single-student distillation from a multi-epoch teacher recovers most of the loss improvement of the ensemble. \citet{busbridge_distillation_2025} establish a teacher--student scaling law that allocates compute optimally between training the teacher and the student. These works model distillation as an internal allocation of compute under a fixed corpus; in our framework, self-distilled tokens are a natural instance of \Dp\ that the same $\eta$ function describes, although our empirical fits in this paper focus on multi-epoch repetition and paraphrasing.
 \paragraph{Data allocation and mixing.}
A complementary line of work allocates compute across the \emph{composition} of the fresh corpus rather than between fresh and derived tokens: DoReMi \citep{xie_doremi_2023}, RegMix \citep{liu_regmix_2025}, and the data-mixing scaling laws of \citet{ye_data_2024} optimize mixture weights across pretraining domains, and mixture dependent scaling laws of \citet{hamidieh2026domain} predict model performance by accounting for data synergy. CD-scaling is orthogonal: we hold the fresh-corpus composition fixed and study how compute should be split between fresh \D\ and derived \Dp.\looseness=-1


\section{Methodology and Experiment Setup}
\subsection{Compute-Data Scaling Law}
\label{sec:cd_formulation}

 The classical Chinchilla scaling law is
 \begin{align}
     \Loss^{\mathrm{Chin}}(\N, \D) \;=\; \E + \frac{\A}{\N^\alpha} + \frac{\B}{\D^\beta},
     \label{eqn:classic_scaling}
 \end{align}
 where \N\ is the number of model parameters and \D\ is the number of fresh pretraining tokens, each seen once. To accommodate synthetic and augmented data, we extend \Cref{eqn:classic_scaling} into a 
\emph{compute--data} (CD) scaling law that shares the constants $(\E, \A, \B, \alpha, \beta)$:
\begin{equation}
    \Loss^{\mathrm{CD}}(\N, \D, \Dp) = \E + \frac{\A}{\N^\alpha}
    + \frac{\B}{\left(\D + \eta^{\strat}\!\left(\N, \TTP, \Dp/\D\right) \cdot \Dp\right)^\beta},
\label{eqn:data_chin}
\end{equation}
 where \Dp\ denotes additional tokens produced from a data-expansion strategy (multi-epoch repetition, paraphrasing, or other forms of synthetic data generation and augmentation methods). The \emph{effectiveness coefficient} 
$\eta \in [0, 1]$ predicts how much one derived token in \Dp\ is worth relative to a fresh token from \D. In \Cref{sec:eta_empirical}, we motivate parameterizing $\eta$ in terms of the three more interpretable quantities $(\N, \TTP, \Dp/\D)$ rather than $(\N, \D, \Dp)$, and in \Cref{sec:eta_form}, we derive a parametric form for $\eta^{\strat}$.

\Cref{eqn:data_chin} has three natural limits. Setting $\Dp = 0$ recovers the classic scaling law $\Loss^{\mathrm{Chin}}$. Taking $\D \to \infty$ sends the third term to zero, leaving the model-bound floor $\Loss^{\N} = \E + \A/\N^\alpha$, the lowest loss a model of size \N\ can reach under unlimited fresh data. Finally, and most importantly, taking $\Dp \to \infty$ at fixed \D\ defines \emph{data-optimal scaling} $\Loss^{\D}$: the lowest loss attainable on a fixed corpus \D\ when unlimited compute is spent on derived tokens. The exact form of $\Loss^{\D}$ depends on $\eta$ and is derived in \Cref{sec:eta_form}.

\subsection{Experiment Setup}
\label{sec:setup}
\paragraph{Data, model, and training.} We use the
OLMo3~\citep{olmo_olmo_2025} architecture and training infrastructure
for all pretraining runs. We train models ranging from $\N = 14$M to
$\N = 600$M parameters, with fresh data \D\ ranging from $30$M to
$30$B tokens and derived tokens \Dp\ from $30$M to $120$B. Fresh
tokens are sampled from the Dolma-3 150B
corpus~\citep{olmo_olmo_2025}\footnote{\url{https://huggingface.co/datasets/allenai/dolma3_mix-150B-1025}},
with smaller \D\ always a strict subset of larger \D. All models are
trained at sequence length 4096, batch size 512, with
AdamW~\citep{adamw} and cosine learning-rate decay. We sweep
learning rate $\in \{1\text{e-}4, 3\text{e-}4, 1\text{e-}3,
3\text{e-}3\}$ and weight decay $\in \{0.1, 0.2, 0.4, 0.8, 1.6\}$
at $\N \in \{30\text{M}, 370\text{M}\}$, selecting the configuration
that minimizes validation loss; for other model sizes we perform a
local grid search around the selected configuration.  Our primary metric is validation loss on a held-out set of 5M Dolma-3 documents. 

\paragraph{Data expansion strategies.}
\label{sec:data_strategies}
We extend pretraining compute beyond 1-epoch training on \D\ via
two data-expansion strategies, each producing additional tokens
\Dp:
\begin{enumerate}[leftmargin=20pt, itemsep=2pt, topsep=-5pt, parsep=0pt]
    \item \textbf{Multi-epoch}~\citep{muennighoff_scaling_2025}:
    repeat the original corpus for additional epochs.
    \item \textbf{Paraphrasing (i.e., augmentation)}~\citep{maini_rephrasing_2024,
    kang_demystifying_2025}: rephrase each document in \D\ using SmolLM2-1.7B-Instruct as the  paraphrasing model~\citep{allal2025smollm2smolgoesbig}, sampling up
    to 16 paraphrases per document with the style for each seed
    drawn uniformly from \{question, math (or wiki when not
    applicable), FAQ, table\}, a four-style mixture identified as
    most effective in prior analysis~\citep{synthetic_data_playbook}.
    Models are then trained on a mixture of fresh \D\ and
    paraphrased \Dp\ for one epoch. Paraphrased documents are always
    derived from documents in \D, and smaller paraphrased corpora
    are nested in larger ones.
\end{enumerate}
We do not include the FLOPs spent on paraphrase generation in
training compute, as paraphrasing can be performed asynchronously
and is not part of the training loop.

\section{Parametric Form of CD-Scaling Law}
\label{sec:scaling_law}

We first characterize $\eta$ empirically (\Cref{sec:eta_empirical}), then we use those observations to motivate a functional form (\Cref{sec:eta_form}), and fit the resulting closed-form (\Cref{sec:fit_results}). We validate that the fit extrapolates to held-out model sizes and show ablations on the parametric form choice (\Cref{sec:ablations}). Finally, we confirm that validation loss improvements transfer to downstream benchmark performance (\Cref{sec:loss_downstream}) across training settings.\looseness=-1

\begin{figure}[t]
    \centering
     \includegraphics[width=1.0\linewidth]{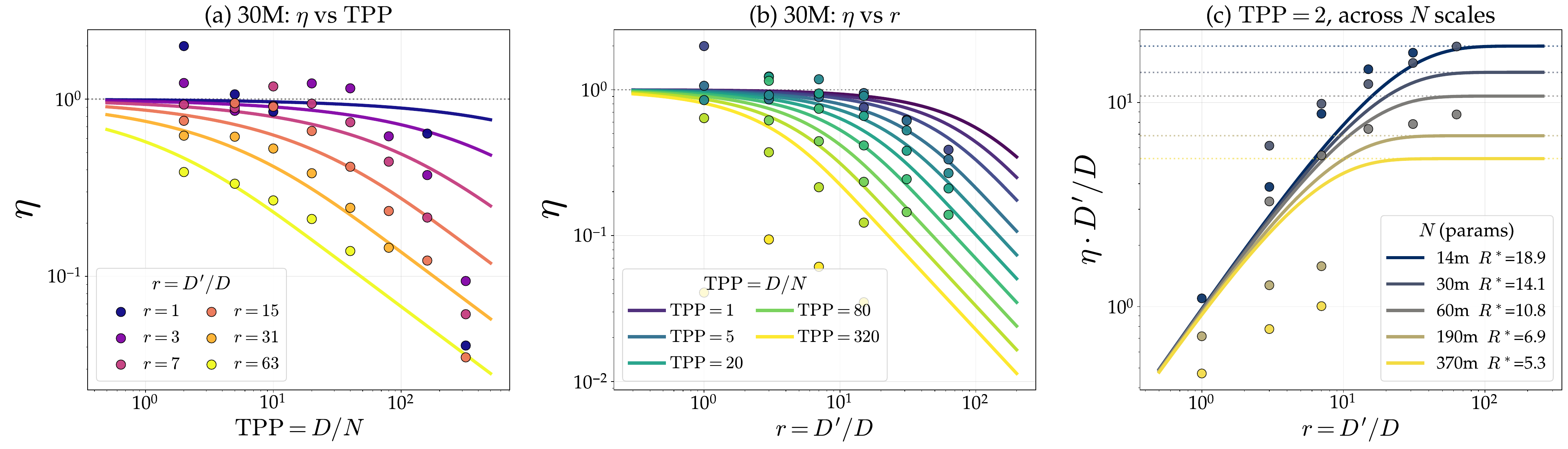}
    \caption{
    \textbf{Empirical characterization of $\eta$} ($y$-axis is $\eta_{\mathrm{emp}}$ throughout).
    \textit{(a)} $\eta$ vs.\ \TTP at fixed $\Dp/\D$: $\eta$ decreases monotonically with \TTP, so the portion of fresh data in the pretraining set matters.
    \textit{(b)} $\eta$ vs.\ $\Dp/\D$ at fixed $(\N, \D)$: monotone decay; together with (c), this is decay \emph{toward a finite saturation ceiling}, not decay to zero.
    \textit{(c)} $\eta \cdot \Dp/\D$ vs.\ $r$ at fixed $(\N, \D)$: extra effective tokens \emph{plateau} as $\Dp$ grows. The plateau is a fundamental property of $\eta$ (and does not arise from under-fitting): no matter how many derived tokens we add, the loss curve flattens, and any functional form for which $\eta \cdot \Dp$ diverges with $\Dp$ is ruled out. We also observe that larger models saturate faster, consistent with findings in \Cref{sec:optimal_scaling_ratio}.
    }    
    \label{fig:eta_analysis}
\end{figure}

\subsection{Empirical Characterization of $\eta$}
\label{sec:eta_empirical}
Before specifying a parametric form for $\eta^{\strat}$, we measure it empirically and examine how it varies with $(\N, \D, \Dp)$. To obtain $\eta_{\mathrm{emp}}$, we first fit the classic scaling law (\Cref{eqn:classic_scaling}) to obtain constants $(\E, \A, \B, \alpha, \beta)$. For each repetition and paraphrasing run, we then define the \emph{empirical} coefficient $\eta_{\mathrm{emp}}$ as the unique value that, substituted into \Cref{eqn:data_chin} alongside these constants, reproduces the observed loss \Loss. We expect $\eta_{\mathrm{emp}} \in [0, 1]$, though it can exceed $1$ when the 1-epoch fit underestimates a run's loss, as any such residual is absorbed into $\eta_{\mathrm{emp}}$. We discuss this limitation in \Cref{sec:conclusion}.

\paragraph{Change of variables.} Although $\eta^{\strat}$ is naturally a function of $(\N, \D, \Dp)$, we recast it in terms of three more interpretable variables: the tokens-per-parameter ratio $\TTP = \D/\N$, the expansion ratio $r = \Dp/\D$, and the model size \N. \TTP\ measures fresh-data availability relative to model capacity, and is the conventional unit in which Chinchilla scaling is described. The expansion ratio $r$ measures how aggressively \Dp expands beyond \D. In the multi-epoch case, $r$ is simply the number of extra epochs. We retain \N\ as a separate variable so that \TTP\ captures data availability and \N\ captures model size alone.

\paragraph{Empirical observations.} In \Cref{fig:eta_analysis}, we examine how $\eta_{\mathrm{emp}}$ depends on $(r, \TTP, \N)$. Panel~\textit{(a)} shows that $\eta_{\mathrm{emp}}$ decays with \TTP, and the decay steepens as $r$ grows. Panel~\textit{(b)} shows that $\eta_{\mathrm{emp}}$ decays with $r$, and the decay rate depends on \TTP: slowly when \TTP\ is small, sharply when \TTP\ is large. The quantity that matters for total loss, however, is not $\eta$ but the effective derived data $\eta \cdot \Dp$, or $\eta \cdot r$ once normalized by \D. This quantity represents the fresh-equivalent tokens the derived corpus $\Dp$ contributes. Rather than growing without bound, $\lim_{r \to \infty} \eta \cdot r$ should saturate. We visualize the saturation behavior in panel~\textit{(c)}. Both the saturation limit and the rate of saturation depend on \N, with larger models saturating at a lower value and at a faster rate.

\subsection{Functional Form of $\eta$}
\label{sec:eta_form}
The empirical observations impose three requirements: (i) $\eta_{\mathrm{emp}} \to 1$ as $r \to 0$, (ii) $\eta \cdot r$ saturates to a finite limit as $r \to \infty$, and (iii) the saturation behavior depends on \TTP\ and \N. Among the functional families satisfying the first two conditions, we adopt the exponential form below, and \Cref{sec:ablations} shows it provides the best fit when compared to all other forms considered:
\begin{align}
    \eta = \frac{R^{*}}{r}\left(1 - e^{-r/R^{*}}\right)
    \label{eq:eta_exp}
\end{align}
\paragraph{Interpretation of $R^{*}$.} The exponential form gives the identity $\lim_{r \to \infty} \eta \cdot r = R^{*}$, so $R^{*}$ can be interpreted as the \emph{saturation ceiling}: the maximum number of fresh-equivalent tokens a corpus of size \D\ can yield through data expansion, expressed as a multiple of \D. This ceiling implies that total effective data saturates at $\lim_{\Dp \to \infty} \D_{\mathrm{eff}} = \D(1 + R^{*})$. Applying this identity to \Cref{eqn:data_chin} yields a closed form for data-optimal scaling $\Loss^{\D}$, the lowest loss a given data-expansion strategy can reach on a fixed corpus of size \D:
\begin{align}
    \Loss^{\D}(\N, \D) \;=\; \E + \frac{\A}{\N^\alpha}
    + \frac{\B}{\left[\D\,\left(1 + R^{*}\left(\D, \N\right)\right)\right]^\beta}.
    \label{eq:data_optimal}
\end{align}
\paragraph{Functional form of $R^{*}$.} Based on observations in \Cref{sec:eta_empirical}, we propose a power law for $R^{*}$:
\begin{align}
    R^{*}(\D, \N) \;=\; \K \cdot (\D/\N)^{\rho} \cdot \N^{\sigma},
    \label{eq:rstar_anchor}
\end{align}
where $\rho$ governs how $R^{*}$ tightens with \TTP\ and $\sigma$ how $R^{*}$ tightens with \N. We overlay the proposed form in \Cref{fig:eta_analysis}, and confirm that the power law fits the observed qualitative behavior. 

Together, \Cref{eq:eta_exp,eq:rstar_anchor} determine the token effectiveness function $\eta$.  Substituting $\eta$ into \Cref{eqn:data_chin} gives the complete CD-scaling law, containing eight constants: the five 1-epoch Chinchilla parameters $(\E, \A, \B, \alpha, \beta)$ and three strategy-specific parameters $(\K, \rho, \sigma)$. The repeated-data law of \citet{muennighoff_scaling_2025} is the special case in which $R^{*}$ is constant and independent of $(\D, \N)$. Ablations in \Cref{sec:ablations} (\Cref{tab:form_ranking}) confirm that this constant-$R^{*}$ assumption fits worse, and that the $(\D, \N)$-dependence in \Cref{eq:rstar_anchor} is what makes the law predictive across model sizes and data-availability regimes.\looseness=-1

\subsection{Fitting CD-Scaling Law}
\label{sec:fit_results}

\begin{table}[t]
\centering
\small
\setlength{\tabcolsep}{5pt}
\vspace{-.2in}
\caption{\textbf{Fitted CD-Scaling parameters with 95\% CIs.}
    \textit{Top:} 1-epoch Chinchilla parameters, 
    \textit{Bottom:} per-strategy CD-Scaling parameters. RMSE on $\log\Loss$, reported separately for 1-epoch, repetition and paraphrasing training runs.}
\label{tab:fit_results}

\begin{tabular}{ccccc c}
\toprule
$\E$ & $\A$ & $\B$ & $\alpha$ & $\beta$ & RMSE$_{\text{1ep}}$ \\
\midrule
\makecell{$1.35$\\[-1pt]\scriptsize$[1.0,\,1.6]$}
 & \makecell{$205$\\[-1pt]\scriptsize$[90,\,676]$}
 & \makecell{$16{,}597$\\[-1pt]\scriptsize$[7{,}677,\,34{,}475]$}
 & \makecell{$0.283$\\[-1pt]\scriptsize$[0.22,\,0.37]$}
 & \makecell{$0.435$\\[-1pt]\scriptsize$[0.40,\,0.47]$}
 & $0.043$ \\
\bottomrule
\end{tabular}

\vspace{6pt}

\begin{tabular}{l ccc c}
\toprule
& $\log \K$ & $\rho$ & $\sigma$ & RMSE \\
\midrule
Repetition
 & \makecell{$10.93$\\[-1pt]\scriptsize$[9.7,\,13.5]$}
 & \makecell{$-0.42$\\[-1pt]\scriptsize$[-0.71,\,-0.15]$}
 & \makecell{$-0.41$\\[-1pt]\scriptsize$[-0.55,\,-0.35]$}
 & $0.035$ \\
Paraphrase
 & \makecell{$30.50$\\[-1pt]\scriptsize$[30.4,\,30.8]$}
 & \makecell{$-1.52$\\[-1pt]\scriptsize$[-1.71,\,-1.22]$}
 & \makecell{$-1.30$\\[-1pt]\scriptsize$[-1.35,\,-1.26]$}
 & $0.024$ \\
\bottomrule
\end{tabular}
\end{table}

\begin{figure}[t]
    \centering
    \includegraphics[width=0.8\linewidth]{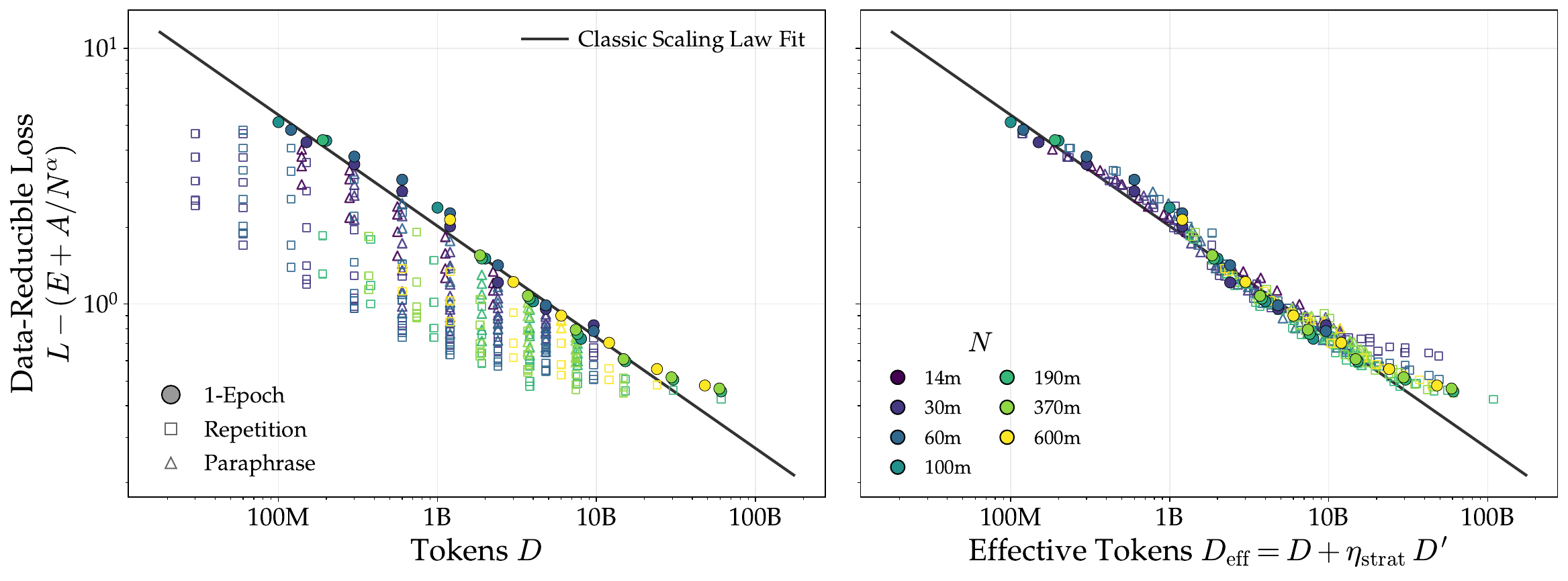}
    \caption{
    \textbf{Fitted CD-scaling law across model sizes and strategies.} Data-reducible loss $\Loss - (\E + \A/\N^{\alpha})$ across model sizes and training settings ($1$-epoch, repetition, and paraphrase). The solid line is the classic scaling-law fit on 1-epoch runs.
    \textit{Left:} Data-reducible loss against \emph{fresh} tokens \D. Repetition and paraphrase runs train on extra $\Dp$ tokens, pushing their loss below what fresh \D\ alone achieves, so these runs fall beneath the classic scaling law.
    \textit{Right:} The CD-scaling law converts each $\Dp$ into its fresh-token equivalent $\eta_{\strat}\Dp$. Re-plotted against the effective token count $D_{\text{eff}} = \D + \eta_{\strat}\Dp$, the paraphrase and repetition runs reduce back to the classic scaling law.
    }
    \label{fig:fitting_result}
\end{figure}

\begin{figure}[t]
    \centering
    \includegraphics[width=0.8\linewidth]{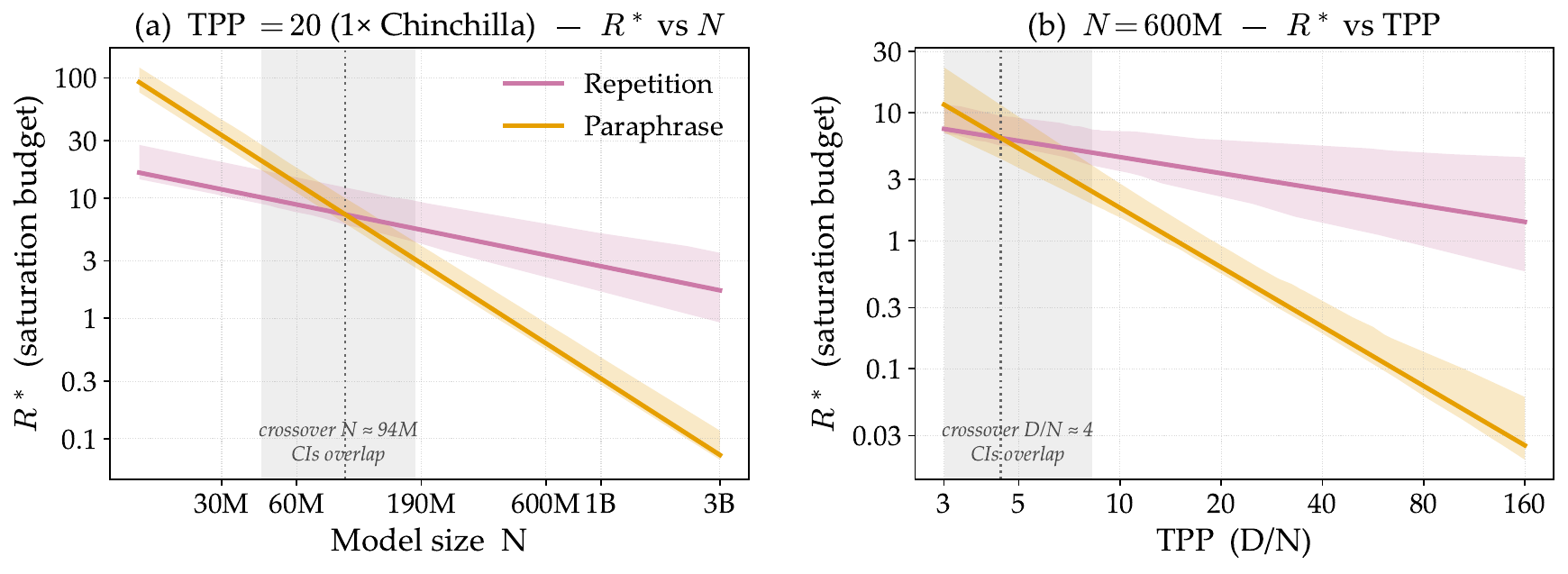}
    \caption{
    \textbf{Saturation ceiling $R^*$ across strategies.} Lines show fitted $R^*$, shaded bands show $95\%$ CIs, and the grey region marks where the two CIs overlap.
    \textit{(a)} $R^*$ vs.\ \N\ at $\TTP=20$. Paraphrasing significantly wins for small models ($\N\lesssim50$M) and repetition for large ones ($\N\gtrsim190$M).
    \textit{(b)} $R^*$ vs.\ \TTP\ for $\N=600$M. The CIs overlap at small \TTP, so paraphrasing does not meaningfully beat repetition. Beyond $\TTP\approx8$, repetition significantly wins.}
    \label{fig:r_star_cross_ci}
    \vspace{-.2in}
\end{figure}

We follow the fitting procedure introduced by \citet{besiroglu2024chinchilla}. Namely, we use a Huber loss on the residuals of $\log \Loss$. We jointly optimize all parameters: the $5$ parameters of the 1-epoch scaling law $(\E, \A, \B, \alpha, \beta)$, together with $3$ per-strategy saturation parameters $(\K, \rho, \sigma)$ for repetition and paraphrasing. We provide the detailed fitting procedure in Appendix~\ref{app:fit} and report the fitted parameters in \Cref{tab:fit_results}.\looseness=-1

To assess fit quality, we separately report the RMSE of $\log \Loss$ for 1-epoch, repetition, and paraphrasing runs. We also use 95\% confidence intervals (CIs) to quantify uncertainty in the fitted parameters. Specifically, we resample training runs with replacement and refit the parameters on the resampled data. We repeat this resampling and refitting procedure $200$ times and take the $95\%$ confidence interval. We report RMSE and per-parameter confidence intervals in \Cref{tab:fit_results} as well.

Fitting the CD-scaling law provides a principled way to compare data-expansion strategies. As an example, we compare saturation ceilings $R^*$ (\Cref{sec:eta_form}). Because we fit CIs on $R^*$, we can go beyond point estimates and ask whether the difference between strategies is statistically significant. In each panel of \Cref{fig:r_star_cross_ci}, the shaded band marks where the two $95\%$ CIs overlap. At the Chinchilla-optimal budget ($\TTP = 20$), \Cref{fig:r_star_cross_ci}~\textit{(a)} compares $R^*$ across model sizes: the repetition and paraphrasing CIs overlap in a band around the crossover ($\N \approx 94$M), but paraphrasing significantly wins for small models ($\N \lesssim 50$M) and repetition for large ones ($\N \gtrsim 190$M). At a fixed size ($\N = 600$M), \Cref{fig:r_star_cross_ci}~\textit{(b)} compares $R^*$ across \TTP: the CIs overlap at small \TTP, so paraphrasing does not meaningfully beat repetition, whereas beyond $\TTP \approx 8$ repetition wins. We map the full crossover boundary across $(\N, \TTP)$ in \Cref{sec:three_regime}.

\subsection{Cross-Scale Validation and Ablations}
\label{sec:ablations}

\paragraph{Cross-scale validation.} A practical motivation for fitting \Cref{eq:eta_exp} is to predict loss for the expensive large-model runs from cheaper small-model runs. We now examine whether the CD-scaling law can extrapolate beyond the model sizes used in fitting. To test this, we refit \Cref{eq:eta_exp} on runs at $\N \in \{14, 30\}$M only, then predict held-out losses at $\N \in \{60, 100, 190, 370, 600\}$M. The held-out set spans model sizes $2\times$ to $20\times$ larger than any used in fitting, \TTP up to ${\sim}160$ ($8\times$ Chinchilla), and $r$ up to $63$ (64 epochs), which covers token-count regimes well outside the fitting range. Additionally, we sweep on model sizes used for fitting in Appendix~\ref{app:xval_sweep}.

The small-$\N$ fit achieves RMSE $0.079$ on $\log \Loss$ on the held-out points, against an in-sample RMSE of $0.048$. Our scaling law therefore transfers cleanly across an order of magnitude in model size, and a practitioner can fit \Cref{eq:eta_exp} on small models with modest budget and read off $R^{*}(\D, \N)$ at production scale to anticipate how much additional data will reduce loss.

\paragraph{Ablations.} We validate the exponential functional form \Cref{eq:eta_exp} and the $R^*$ form \Cref{eq:rstar_anchor} using multi-epoch training runs, reporting leave-one-out (LOO) RMSE on $\log \Loss$ under the same fitting procedure.\looseness=-1

For the functional form of $\eta$, we compare nine candidates (Appendix~\ref{app:eta_forms}, \Cref{tab:form_ranking}). The forms that let $R^{*}$ depend on $(\D, \N)$ via \Cref{eq:rstar_anchor} dominate the rest at every parameter count; in particular, the constant-$R^{*}$ data-repetition baseline \citep{muennighoff_scaling_2025} is rejected by ${\sim}40\%$ in relative LOO RMSE. Within the top group, our chosen form is statistically indistinguishable from the best, and we prefer it because it generalizes the constant-$R^{*}$ form and satisfies $\eta(0) = 1$ exactly.

For $R^{*}$, we remove terms from \Cref{eq:rstar_anchor} and refit (Appendix~\ref{app:term_ablation}, \Cref{tab:term_ablation}). Dropping either exponent degrades the fit, and dropping both recovers the constant-$R^{*}$ baseline: each exponent is individually necessary, and using both is strictly better than either alone.

\subsection{Loss-to-Downstream Transfer}
\label{sec:loss_downstream}

\begin{figure}[t]
    \centering
    \includegraphics[width=0.95\linewidth]{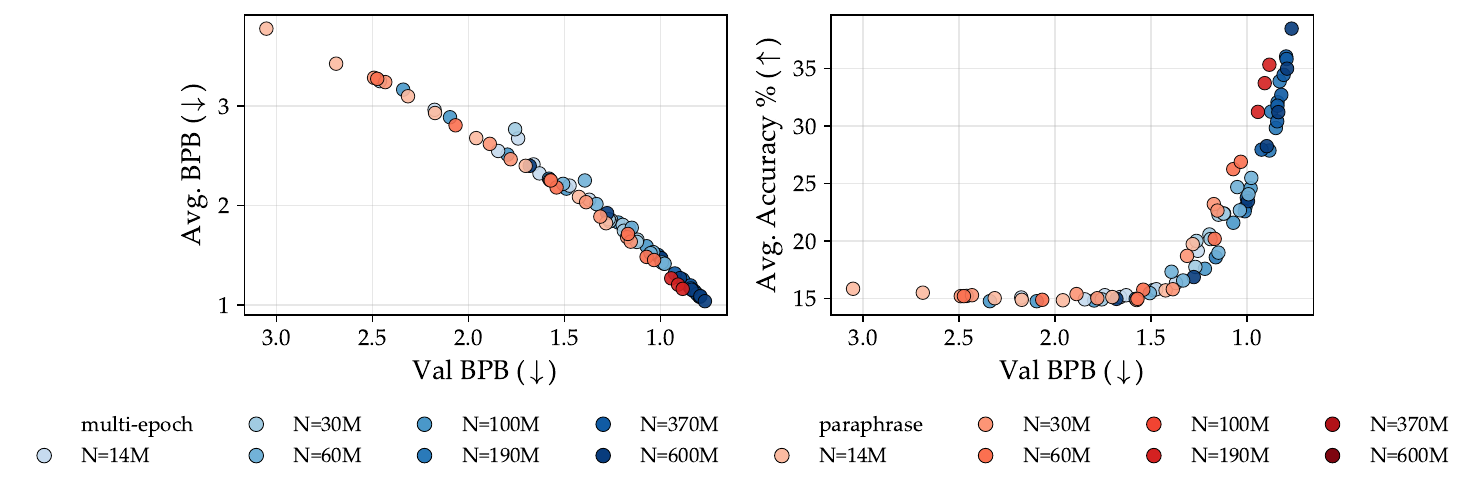}
    \caption{
    \textbf{Downstream performance as a function of validation BPB.}
    \textit{Left}: average BPB on benchmarks recast as language-modeling tasks (GSM8K, TriviaQA, HumanEval, etc.). \textit{Right}: mean accuracy on LAMBADA, HellaSwag, OpenBookQA, RACE, and SQuAD. In both panels, points from multi-epoch and paraphrase runs collapse onto a single curve, indicating that validation loss is a sufficient statistic for downstream capability regardless of training settings.
    }
    \label{fig:loss_benchmark_main}
\end{figure}

So far we have compared validation loss across training settings. We now ask whether validation loss remains a valid predictor of downstream capability across data-expansion strategies and model scales.

We group benchmarks into LM-based and accuracy-based tasks. Benchmarks such as GSM8K~\citep{cobbe2021training}, TriviaQA~\citep{joshi2017triviaqa}, and HumanEval~\citep{chen2021evaluating} yield near-random accuracy at small-to-medium scale. Following~\citet{gadre_language_2024}, we recast them as language-modeling tasks and report bits-per-byte (BPB) of the gold response given the prompt, which stays predictive even when accuracy is near-trivial. For accuracy-based tasks, we report mean accuracy across LAMBADA~\citep{paperno2016lambada}, HellaSwag~\citep{zellers2019hellaswag}, OpenBookQA~\citep{OpenBookQA2018}, RACE~\citep{lai2017race}, and SQuAD~\citep{rajpurkar-etal-2018-know}. We plot both against validation BPB in \Cref{fig:loss_benchmark_main}.

In both cases, runs from all strategies and scales fall onto a single curve as a function of validation BPB alone, which confirms that validation loss predicts downstream capability independent of training setting. We report per-task breakdowns in \Cref{app:loss_benchmark_breakdown}.

\section{CD-Scaling Laws Bridge Compute- and Data-Optimal Training}
\label{sec:analysis}
In this section, we discuss three implications of CD-scaling laws. First, for a fixed \N, the CD law predicts the compute-data Pareto allocation between \D\ and \Dp\ (\Cref{sec:optimal_scaling_ratio}). Second, it partitions training into three regimes, identifying which resource is the binding constraint for a given $(\N, \D, \Dp)$ (\Cref{sec:three_regime}). Third, it tells a practitioner which data-expansion strategy to adopt and how much compute to spend before returns saturate (\Cref{sec:practical_implication}).

\begin{figure}[t]
    \centering
    \includegraphics[width=1.0\linewidth]{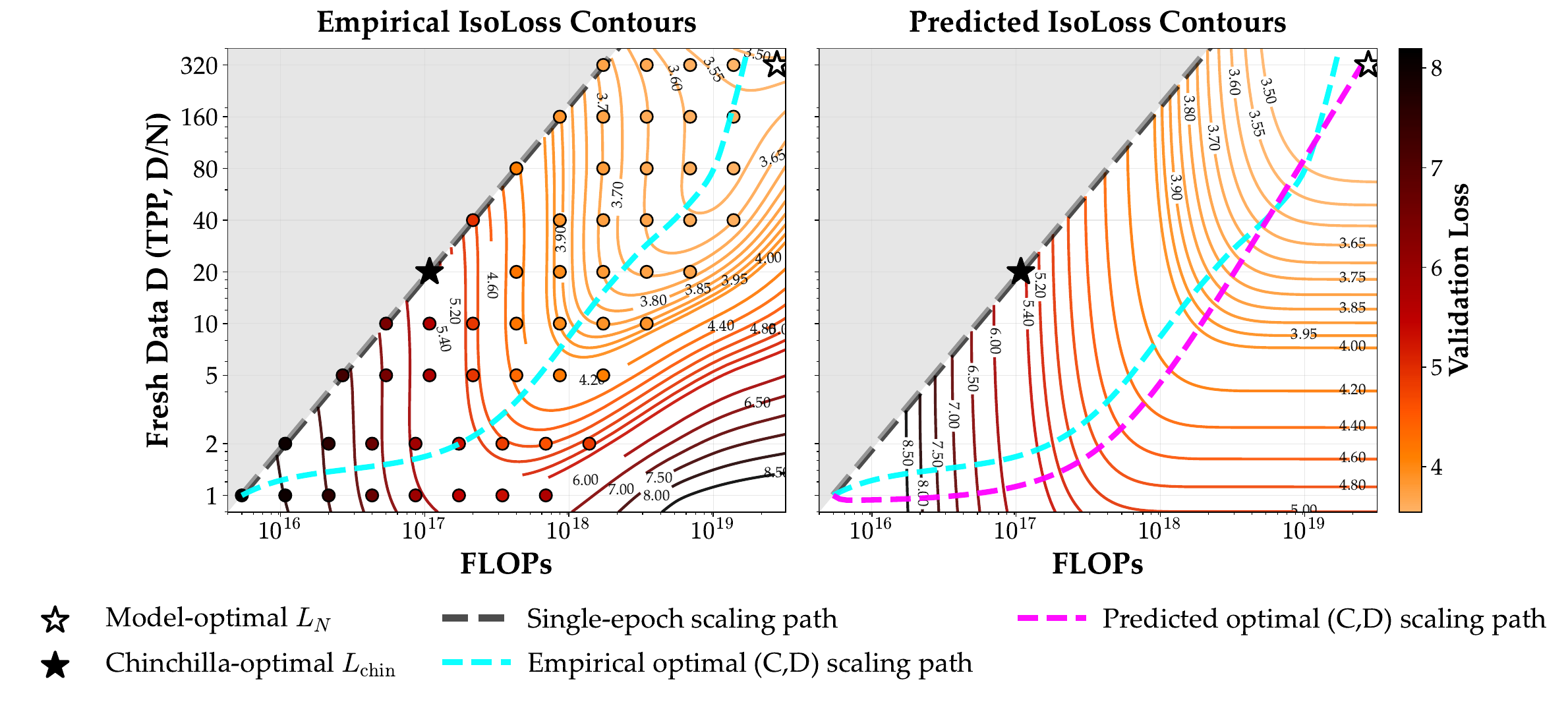}
    \caption{
    \textbf{Optimal compute-data allocation on (\C, \D) loss surfaces. }
    \textit{Left:} 1-epoch compute-optimal training scales compute \C proportionally with fresh data \D, reducing the (\C, \D) loss surface into a single ray (\textit{black)}. Loss contour lines around it indicates that small increase in \C would significantly lower the loss. 
    Tracing the sharpest direction of descent, the empirical optimal scaling path (\textit{teal}) prefers scaling compute at data-scarce regime, then scaling both data and compute. 
    \textit{Right:} Our proposed CD-scaling law predicts the optimal path that jointly optimizes (\C, \D).
    }
    \label{fig:multi_epoch_scaling}
\end{figure}

\subsection{Compute-Data Pareto-Optimal Allocation}
\label{sec:optimal_scaling_ratio}

A practitioner training a model of size \N\ on a fixed corpus \D\ must decide how much compute to spend on further loss reduction. Under Chinchilla scaling, compute is fixed by $(\N, \D)$ via $C \approx 6\N\D$, leaving no such freedom. Decoupling \C\ from \D\ through multi-epoch training or paraphrasing adds a degree of freedom, giving a two-dimensional loss surface over $(\C, \D)$ with $C \approx 6\N(\D + \Dp)$. \Cref{fig:multi_epoch_scaling}~\textit{(left)} shows this empirical surface for $\N = 30$M, with compute \C\ on the $x$-axis and fresh data \D\ on the $y$-axis. The black dashed line traces single-epoch runs, where the surface collapses onto a single ray. The contours around single-epoch training are far steeper in \C\ than in \D, indicating that modest compute beyond the single-epoch buys substantial loss reductions that single-epoch scaling leaves on the table.\looseness=-1

\paragraph{The empirical Pareto frontier is non-trivial.} The teal curve traces the loss-minimizing path through $(\C, \D)$ space via steepest descent on the empirical surface. Along it, the contours bend sharply in both directions, meaning increasing \C\ or \D\ \emph{alone} would leave loss roughly unchanged. Since neither resource can reduce loss in isolation, this line traces the compute-data Pareto frontier. This frontier gives practitioners a general guidance on resource allocation. In the data-scarce, low-compute regime (bottom-left), increasing compute is far more effective than adding fresh data, even when data is scarce. As \C\ grows, the frontier rotates toward scaling \C\ and \D\ in equal proportion, then enters a basin where fresh data dominates and further compute yields diminishing returns.

\paragraph{CD-scaling predicts this frontier.} Tracing the steepest-descent path on the CD-scaling loss surface (\Cref{fig:multi_epoch_scaling}~\textit{right}) yields a trajectory that closely tracks the empirical Pareto curve, generalizing Chinchilla's compute-only optimization into a joint compute-data optimization.

\subsection{Three Training Regimes}
\label{sec:three_regime}

\begin{figure}[t]
    \centering
    \includegraphics[width=1.0\linewidth]{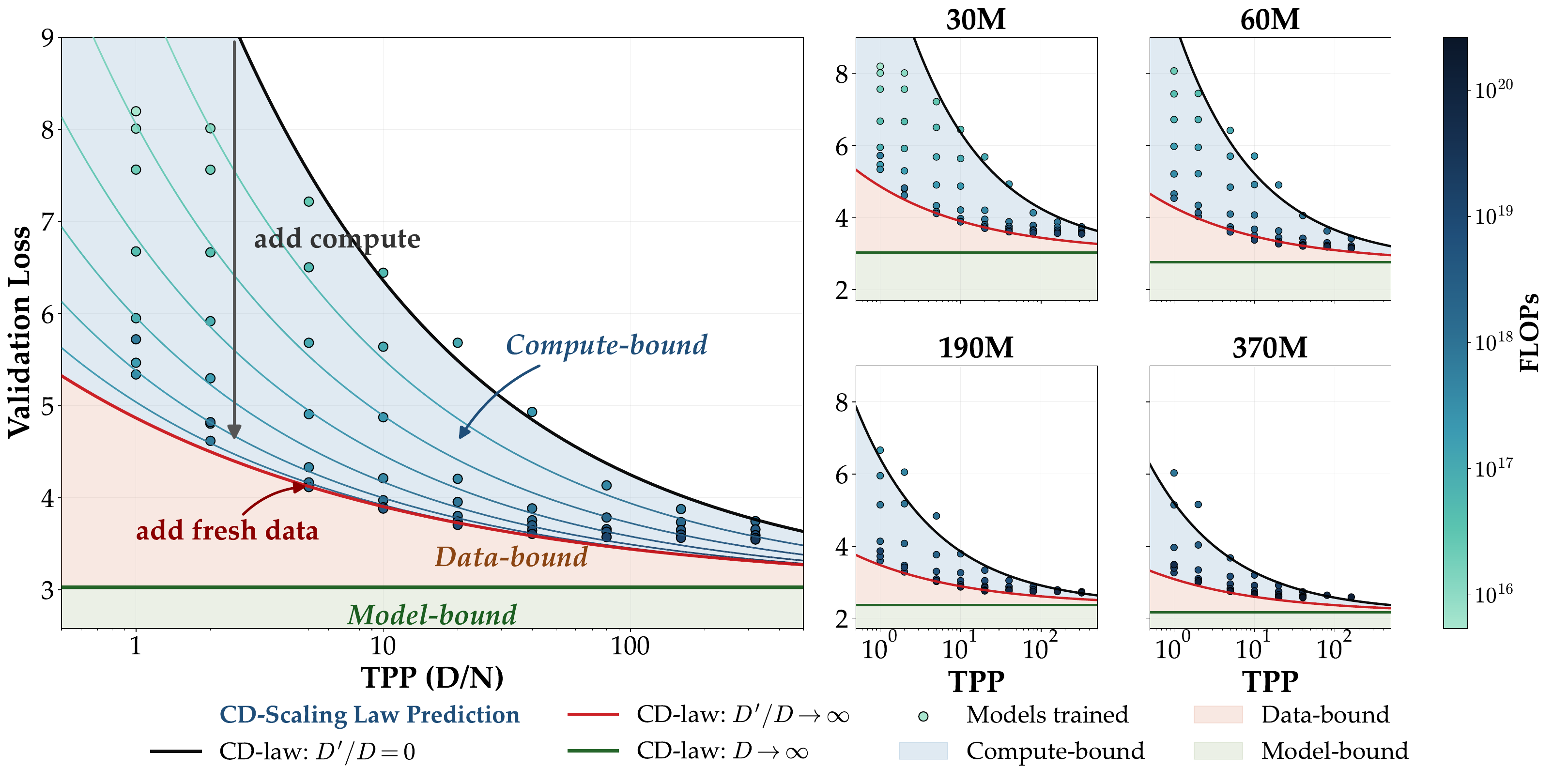}
    \caption{
    \textbf{The CD-scaling law captures three training regimes.}
    \textit{Left}: $N=30$M empirical losses overlaid with CD-scaling law predictions. By sweeping \Dp from $0$ ($\Loss^{\mathrm{Chin}}$, compute-optimal limit) to $\infty$ ($\Loss^{\D}$, data-optimal limit), CD-scaling law traces out all achievable losses from $\Loss^{\mathrm{Chin}}$ to $\Loss^{\D}$. 
    \textit{Right}: The same decomposition across model sizes $\N$ . As \N grows, the compute-bound region bounded by $\Loss^{\mathrm{Chin}}$ and $\Loss^{\D}$ tightens, indicating that for larger models, increasing compute \C yields diminishing returns in further loss reduction. 
}
    \label{fig:data_scaling_law}
\end{figure}

As fresh data become the binding constraint, instead of plotting validation loss against training compute, we plot against fresh data, with color indicating the compute budget $\C \approx 6\N(\D + \Dp)$, shown in \Cref{fig:data_scaling_law}. The three limits of CD-scaling: $\Loss^{\mathrm{Chin}}$ ($\Dp = 0$), $\Loss^{\D}$ ($\Dp \to \infty$), and $\Loss^{\N}$ ($\D \to \infty$) appear as three boundary curves that carve the (\Loss, \D) plane into three distinct regions:
\begin{itemize}[itemsep=0pt,labelindent=2pt,topsep=0pt,parsep=0pt,partopsep=1pt,align=left,leftmargin=*]
    \item \textbf{Compute-bound} (between $\Loss^{\mathrm{Chin}}$ and
    $\Loss^{\D}$): for a given \D, additional $\Dp$ still reduces loss.
    Compute is the binding resource.
    \item \textbf{Data-bound} (between $\Loss^{\D}$ and $\Loss^{\N}$):
    $\Dp$ has saturated to $R^*\D$, only additional data can further reduce loss.\looseness=-1
    \item \textbf{Model-bound} (at $\Loss^{\N}$): both compute and data
    have saturated. \N\ is the binding resource.
\end{itemize}

We then overlay the three region boundaries predicted by CD-law against empirical training losses. Single-epoch points lie on $\Loss^{\mathrm{Chin}}$. Intermediate-$\Dp$ traces countour lines in the compute-bound region. Importantly, the predicted data-optimal frontier $\Loss^{\D}$ coincides with the observed $\Dp \to \infty$ asymptote of the training runs, validating that the CD-scaling's saturation limit correctly models the empirical limit.\looseness=-1

\paragraph{The compute-bound region shrinks with both \TTP\ and \N.} The vertical gap between $\Loss^{\mathrm{Chin}}$ and $\Loss^{\D}$ is governed by the saturation ceiling $R^{*}$. When $R^{*}$ is large, the data-optimal frontier sits well below the compute-optimal one. In this training regime, additional training compute spent on $\Dp$ delivers substantial loss reductions. In contrast, when $R^{*}$ is small, the two frontiers nearly coincide and additional compute spent on $\Dp$ is largely ineffective.
In \Cref{sec:eta_empirical}, we empirically observed that $R^*$ shrinks with both \N and \TTP.
In \Cref{fig:data_scaling_law}, we confirm that the fitted CD-scaling law correctly models the shrinking rate over $\TTP$~\textit{(left)}, and $N$~\textit{(right)}.

\subsection{Practical Implications}
\label{sec:practical_implication}

\begin{figure}[h]
    \centering
     \includegraphics[width=1.0\linewidth]{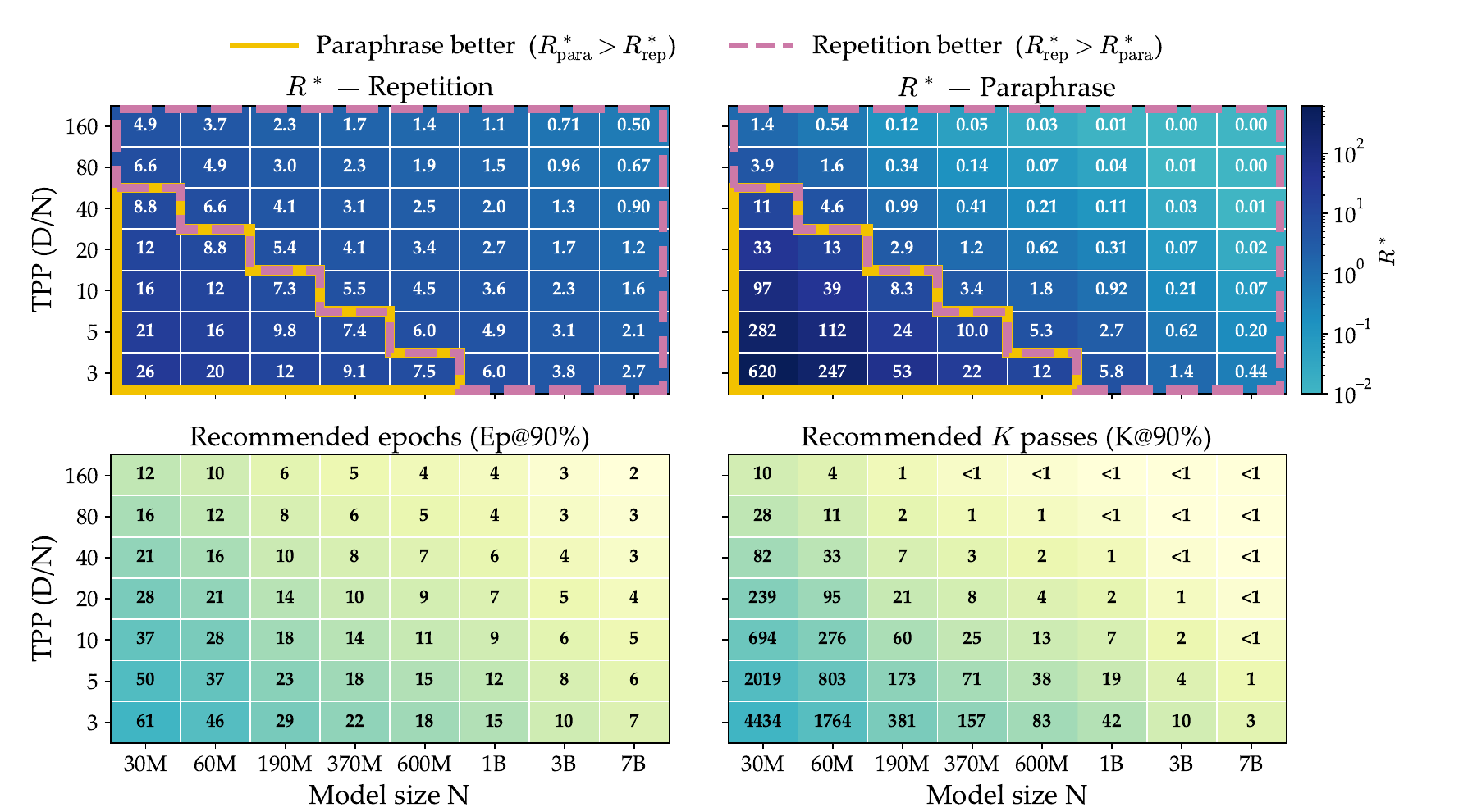}
        \caption{\textbf{When to repeat, when to paraphrase, and how much.} 
            \textit{Top:} Saturation ceiling $R^*$ across $\N$ and $\TTP$ for repetition (\textit{left}) and paraphrase (\textit{right}). The boundary marks where the winning strategy switches. Paraphrase wins in the lower-left (small $\N$, low data budget), repetition wins in the upper-right. 
            \textit{Bottom:} Training compute to reach $90\%$ of $R^*$. We provide recommended epochs (\textit{left}) for repetition and passes $K$ (\textit{right}) for paraphrase. Both decrease as $\N$ and $\TTP$ grow.}    
            \label{fig:practical_heatmap}
\end{figure}

By visualizing saturation ceiling $R^*$ predicted by CD-scaling laws, we can directly understand at different $\N$ and $\TTP$, which of the two data expansion  strategy to adopt and the best achievable outcome, as shown in 
\Cref{fig:practical_heatmap}~(\textit{top}). \Cref{fig:practical_heatmap}~(\textit{bottom}) then reports the training effort needed to reach $90\%$ of such $R^*$'s. Namely, we report the total epochs for repetition runs and total paraphrasing passes for paraphrase runs. We provide three takeaways:
\begin{itemize}[itemsep=0pt,labelindent=2pt,topsep=0pt,parsep=0pt,partopsep=1pt,align=left,leftmargin=*]
\item The recommended epoch count \textbf{decreases} as model size and data budget \textbf{grow}. Extrapolating the fitted law to large model scales indicates that 4-epoch training is recommended only for medium-scale models ($\sim 3$B) around $1\times$ Chinchilla data scale.
\item For smaller models and data budgets, paraphrasing is preferred. Concretely, paraphrasing beats repetition for models $\lesssim 600$M and small data budgets ($< 1\times$ Chinchilla).
\item Extrapolating the fitted law to large model scales indicates that paraphrasing becomes ineffective for $N\geq 7$B and large data budgets ($\geq 4\times$ Chinchilla).
\end{itemize}
\section{Conclusion \& Discussion}
\label{sec:conclusion}
We establish a unified framework that bridges compute- and data-optimal training. Through the effectiveness function $\eta^{\strat}$, we characterize how much additional compute can reduce loss on a fixed pretraining corpus \D. Building on it, we derive three practical implications: the compute-data Pareto-optimal allocation between \D\ and \Dp, the three-regime picture (compute-bound, data-bound, and model-bound) that identifies the binding constraint, and guidance on which expansion strategy to adopt and how much compute to spend before returns saturate. Together these give practitioners a principled basis for deciding when to invest in compute, fresh data, or model capacity.

\paragraph{Predicting $\eta$ from corpus statistics.} We currently fit $\eta^{\strat}$ empirically per strategy. A natural extension is to predict $\eta$ from dataset statistics alone, using summary statistics such as token diversity, $n$-gram overlap, or perplexity distributions. Such a predictive theory would let practitioners assess an expansion strategy on a given corpus before committing compute, making the framework substantially more practical.\looseness=-1

\paragraph{Limitations.} The functional form we adopt has a clean interpretation, but $\eta$ absorbs any residual misspecification of the 1-epoch fit (and can push $\eta_{\mathrm{emp}}$ above 1), so the overall fit is sensitive to the quality of the 1-epoch runs. Within each run, we swept only learning rate and weight decay. Other hyperparameters (batch size, sequence length, schedule) were held fixed and could in principle interact with $\eta$. Establishing these laws incurs substantiation computation cost. To complete training runs reported in this study, we have consumed over 250k H100-hours. Even with this compute budget, we could only fit scaling laws for two data expansion strategy on small-to-medium scale model sizes. Finally, we evaluate two data-expansion strategies (repetition and paraphrase); others such as self-distillation or synthetic structured data~\citep{lee2026traininglanguagemodelsneural} remain unexplored, as do combinations of strategies that may yield higher effective $\eta$ than any single one.

\section*{Acknowledgments}
DAM acknowledges support from the Kempner Institute, FAS Dean's Competitive Fund for Promising Scholarship, Aramont Fellowship Fund, and the NSF AI-SDM Institute (Grant No. IIS-2229881). Additionally, we thank Antonio Torralba for helpful discussions and guidance on this project.

\clearpage
\bibliographystyle{plainnat}
\bibliography{references, references-2}

@misc{adamw,
      title={Decoupled Weight Decay Regularization}, 
      author={Ilya Loshchilov and Frank Hutter},
      year={2019},
      eprint={1711.05101},
      archivePrefix={arXiv},
      primaryClass={cs.LG},
      url={https://arxiv.org/abs/1711.05101}, 
}

@InProceedings{villalobos2024will,
  title = 	 {Position: Will we run out of data? {L}imits of {LLM} scaling based on human-generated data},
  author =       {Villalobos, Pablo and Ho, Anson and Sevilla, Jaime and Besiroglu, Tamay and Heim, Lennart and Hobbhahn, Marius},
  booktitle = 	 {Proceedings of the 41st International Conference on Machine Learning},
  pages = 	 {49523--49544},
  year = 	 {2024},
  editor = 	 {Salakhutdinov, Ruslan and Kolter, Zico and Heller, Katherine and Weller, Adrian and Oliver, Nuria and Scarlett, Jonathan and Berkenkamp, Felix},
  volume = 	 {235},
  series = 	 {Proceedings of Machine Learning Research},
  month = 	 {21--27 Jul},
  publisher =    {PMLR},
  url = 	 {https://proceedings.mlr.press/v235/villalobos24a.html}
}

@article{sevilla2022compute,
  title={Compute Trends Across Three Eras of Machine Learning},
  author={Sevilla, Jaime and Heim, Lennart and Ho, Anson and Besiroglu, Tamay and Hobbhahn, Marius and Villalobos, Pablo},
  journal={arXiv preprint arXiv:2202.05924},
  year={2022},
  url={https://arxiv.org/abs/2202.05924}
}

@misc{gadre_language_2024,
   title        = {Language Models Scale Reliably with Over-Training and on Downstream Tasks},
   author       = {Gadre, Samir Yitzhak and Smyrnis, Georgios and Shankar, Vaishaal and Gururangan, Suchin and Wortsman, Mitchell and Shao, Rulin and Mercat, Jean and Fang, Alex and Li, Jeffrey and Keh, 
Sedrick and Xin, Rui and Nezhurina, Marianna and Vasiljevic, Igor and Jitsev, Jenia and Soldaini, Luca and Dimakis, Alexandros G. and Ilharco, Gabriel and Koh, Pang Wei and Song, Shuran and Kollar, Thomas and 
Carmon, Yair and Dave, Achal and Heckel, Reinhard and Muennighoff, Niklas and Schmidt, Ludwig},
   year         = {2024},
   eprint       = {2403.08540},
   archivePrefix= {arXiv},
   primaryClass = {cs.CL},
   url          = {https://arxiv.org/abs/2403.08540}
 }

@inproceedings{xie_doremi_2023,
   title     = {{DoReMi}: Optimizing Data Mixtures Speeds Up Language Model Pretraining},
   author    = {Xie, Sang Michael and Pham, Hieu and Dong, Xinyun and Du, Nan and Liu, Hanxiao and Lu, Yifeng and Liang, Percy S. and Le, Quoc V. and Ma, Tengyu and Yu, Adams Wei},
   booktitle = {Advances in Neural Information Processing Systems (NeurIPS)},
   year      = {2023},
   url       = {https://arxiv.org/abs/2305.10429}
 }

@inproceedings{liu_regmix_2025,
   title     = {{RegMix}: Data Mixture as Regression for Language Model Pre-training},
   author    = {Liu, Qian and Zheng, Xiaosen and Muennighoff, Niklas and Zeng, Guangtao and Dou, Longxu and Pang, Tianyu and Jiang, Jing and Lin, Min},
   booktitle = {International Conference on Learning Representations (ICLR)},
   year      = {2025},
   url       = {https://arxiv.org/abs/2407.01492}
 }

@misc{ye_data_2024,
   title        = {Data Mixing Laws: Optimizing Data Mixtures by Predicting Language Modeling Performance},
   author       = {Ye, Jiasheng and Liu, Peiju and Sun, Tianxiang and Zhou, Yunhua and Zhan, Yuhao and Qiu, Xipeng},
   year         = {2024},
   eprint       = {2403.16952},
   archivePrefix= {arXiv},
   primaryClass = {cs.CL},
   url          = {https://arxiv.org/abs/2403.16952}
 }

@article{hamidieh2026domain,
  title={Domain-aware scaling laws uncover data synergy},
  author={Hamidieh, Kimia and Mackey, Lester and Alvarez-Melis, David},
  journal={arXiv preprint arXiv:2607.11052},
  year={2026}
}

@article{besiroglu2024chinchilla,
  title={Chinchilla scaling: A replication attempt},
  author={Besiroglu, Tamay and Erdil, Ege and Barnett, Matthew and You, Josh},
  journal={arXiv preprint arXiv:2404.10102},
  year={2024}
}

@misc{allal2025smollm2smolgoesbig,
      title={SmolLM2: When Smol Goes Big -- Data-Centric Training of a Small Language Model}, 
      author={Loubna Ben Allal and Anton Lozhkov and Elie Bakouch and Gabriel Martín Blázquez and Guilherme Penedo and Lewis Tunstall and Andrés Marafioti and Hynek Kydlíček and Agustín Piqueres Lajarín and Vaibhav Srivastav and Joshua Lochner and Caleb Fahlgren and Xuan-Son Nguyen and Clémentine Fourrier and Ben Burtenshaw and Hugo Larcher and Haojun Zhao and Cyril Zakka and Mathieu Morlon and Colin Raffel and Leandro von Werra and Thomas Wolf},
      year={2025},
      eprint={2502.02737},
      archivePrefix={arXiv},
      primaryClass={cs.CL},
      url={https://arxiv.org/abs/2502.02737}, 
}

@misc{synthetic_data_playbook,
  title={The Synthetic Data Playbook: Generating Trillions of the Finest Tokens},
  author={Joel Niklaus and Guilherme Penedo and Hynek Kydlicek and Elie Bakouch and Lewis Tunstall and Ed Beeching and Thibaud Frere and Colin Raffel and Leandro von Werra and Thomas Wolf},
  year={2026},
  
}

@misc{lee2026traininglanguagemodelsneural,
      title={Training Language Models via Neural Cellular Automata}, 
      author={Dan Lee and Seungwook Han and Akarsh Kumar and Pulkit Agrawal},
      year={2026},
      eprint={2603.10055},
      archivePrefix={arXiv},
      primaryClass={cs.LG},
      url={https://arxiv.org/abs/2603.10055}, 
}

@article{paperno2016lambada,
  title={The LAMBADA dataset: Word prediction requiring a broad discourse context},
  author={Paperno, Denis and Kruszewski, Germ{\'a}n and Lazaridou, Angeliki and Pham, Quan Ngoc and Bernardi, Raffaella and Pezzelle, Sandro and Baroni, Marco and Boleda, Gemma and Fern{\'a}ndez, Raquel},
  journal={arXiv preprint arXiv:1606.06031},
  year={2016}
}

@article{zellers2019hellaswag,
  title={Hellaswag: Can a machine really finish your sentence?},
  author={Zellers, Rowan and Holtzman, Ari and Bisk, Yonatan and Farhadi, Ali and Choi, Yejin},
  journal={arXiv preprint arXiv:1905.07830},
  year={2019}
}

@article{chen2021evaluating,
  title={Evaluating large language models trained on code},
  author={Chen, Mark and Tworek, Jerry and Jun, Heewoo and Yuan, Qiming and Pinto, Henrique Ponde De Oliveira and Kaplan, Jared and Edwards, Harri and Burda, Yuri and Joseph, Nicholas and Brockman, Greg and others},
  journal={arXiv preprint arXiv:2107.03374},
  year={2021}
}

@article{cobbe2021training,
  title={Training verifiers to solve math word problems},
  author={Cobbe, Karl and Kosaraju, Vineet and Bavarian, Mohammad and Chen, Mark and Jun, Heewoo and Kaiser, Lukasz and Plappert, Matthias and Tworek, Jerry and Hilton, Jacob and Nakano, Reiichiro and others},
  journal={arXiv preprint arXiv:2110.14168},
  year={2021}
}

@inproceedings{lai2017race,
  title={Race: Large-scale reading comprehension dataset from examinations},
  author={Lai, Guokun and Xie, Qizhe and Liu, Hanxiao and Yang, Yiming and Hovy, Eduard},
  booktitle={Proceedings of the 2017 conference on empirical methods in natural language processing},
  pages={785--794},
  year={2017}
}

@inproceedings{joshi2017triviaqa,
  title={Triviaqa: A large scale distantly supervised challenge dataset for reading comprehension},
  author={Joshi, Mandar and Choi, Eunsol and Weld, Daniel S and Zettlemoyer, Luke},
  booktitle={Proceedings of the 55th Annual Meeting of the Association for Computational Linguistics (Volume 1: Long Papers)},
  pages={1601--1611},
  year={2017}
}

@inproceedings{rajpurkar-etal-2018-know,
    title = "Know What You Don{'}t Know: Unanswerable Questions for {SQ}u{AD}",
    author = "Rajpurkar, Pranav  and
      Jia, Robin  and
      Liang, Percy",
    editor = "Gurevych, Iryna  and
      Miyao, Yusuke",
    booktitle = "Proceedings of the 56th Annual Meeting of the Association for Computational Linguistics (Volume 2: Short Papers)",
    month = jul,
    year = "2018",
    address = "Melbourne, Australia",
    publisher = "Association for Computational Linguistics",
    url = "https://aclanthology.org/P18-2124",
    doi = "10.18653/v1/P18-2124",
    pages = "784--789",
    eprint={1806.03822},
    archivePrefix={arXiv},
    primaryClass={cs.CL}
}

@inproceedings{OpenBookQA2018,
 title={Can a Suit of Armor Conduct Electricity? A New Dataset for Open Book Question Answering},
 author={Todor Mihaylov and Peter Clark and Tushar Khot and Ashish Sabharwal},
 booktitle={EMNLP},
 year={2018}
}

@misc{lovelace2026prescriptivescalinglawsdata,
      title={Prescriptive Scaling Laws for Data Constrained Training}, 
      author={Justin Lovelace and Christian Belardi and Srivatsa Kundurthy and Shriya Sudhakar and Kilian Q. Weinberger},
      year={2026},
      eprint={2605.01640},
      archivePrefix={arXiv},
      primaryClass={cs.LG},
      url={https://arxiv.org/abs/2605.01640}, 
}

@misc{yan_larger_2025,
	title = {Larger {Datasets} {Can} {Be} {Repeated} {More}: {A} {Theoretical} {Analysis} of {Multi}-{Epoch} {Scaling} in {Linear} {Regression}},
	shorttitle = {Larger {Datasets} {Can} {Be} {Repeated} {More}},
	url = {http://arxiv.org/abs/2511.13421},
	doi = {10.48550/arXiv.2511.13421},
	urldate = {2026-02-09},
	publisher = {arXiv},
	author = {Yan, Tingkai and Wen, Haodong and Li, Binghui and Luo, Kairong and Chen, Wenguang and Lyu, Kaifeng},
	month = nov,
	year = {2025},
	note = {arXiv:2511.13421 [cs]},
}

@misc{busbridge_distillation_2025,
	title = {Distillation {Scaling} {Laws}},
	url = {http://arxiv.org/abs/2502.08606},
	doi = {10.48550/arXiv.2502.08606},
	urldate = {2026-03-23},
	publisher = {arXiv},
	author = {Busbridge, Dan and Shidani, Amitis and Weers, Floris and Ramapuram, Jason and Littwin, Etai and Webb, Russ},
	month = jul,
	year = {2025},
	note = {arXiv:2502.08606 [cs]},
}

@misc{hinton_distilling_2015,
	title = {Distilling the {Knowledge} in a {Neural} {Network}},
	url = {http://arxiv.org/abs/1503.02531},
	doi = {10.48550/arXiv.1503.02531},
	urldate = {2026-03-23},
	publisher = {arXiv},
	author = {Hinton, Geoffrey and Vinyals, Oriol and Dean, Jeff},
	month = mar,
	year = {2015},
	note = {arXiv:1503.02531 [stat]},
}

@misc{yang_synthetic_2025,
	title = {Synthetic bootstrapped pretraining},
	url = {http://arxiv.org/abs/2509.15248},
	doi = {10.48550/arXiv.2509.15248},
	urldate = {2026-03-23},
	publisher = {arXiv},
	author = {Yang, Zitong and Zhang, Aonan and Liu, Hong and Hashimoto, Tatsunori and Candès, Emmanuel and Wang, Chong and Pang, Ruoming},
	month = dec,
	year = {2025},
	note = {arXiv:2509.15248 [cs]},
}

@misc{askari-hemmat_improving_2025,
	title = {Improving the {Scaling} {Laws} of {Synthetic} {Data} with {Deliberate} {Practice}},
	url = {http://arxiv.org/abs/2502.15588},
	doi = {10.48550/arXiv.2502.15588},
	urldate = {2026-03-23},
	publisher = {arXiv},
	author = {Askari-Hemmat, Reyhane and Pezeshki, Mohammad and Dohmatob, Elvis and Bordes, Florian and Astolfi, Pietro and Hall, Melissa and Verbeek, Jakob and Drozdzal, Michal and Romero-Soriano, Adriana},
	month = feb,
	year = {2025},
	note = {arXiv:2502.15588 [cs]},
}

@misc{maini_rephrasing_2024,
	title = {Rephrasing the {Web}: {A} {Recipe} for {Compute} and {Data}-{Efficient} {Language} {Modeling}},
	shorttitle = {Rephrasing the {Web}},
	url = {http://arxiv.org/abs/2401.16380},
	doi = {10.48550/arXiv.2401.16380},
	urldate = {2026-03-23},
	publisher = {arXiv},
	author = {Maini, Pratyush and Seto, Skyler and Bai, He and Grangier, David and Zhang, Yizhe and Jaitly, Navdeep},
	month = jan,
	year = {2024},
	note = {arXiv:2401.16380 [cs]},
}

@misc{kim_pre-training_2025,
	title = {Pre-training under infinite compute},
	url = {http://arxiv.org/abs/2509.14786},
	doi = {10.48550/arXiv.2509.14786},
	urldate = {2026-02-09},
	publisher = {arXiv},
	author = {Kim, Konwoo and Kotha, Suhas and Liang, Percy and Hashimoto, Tatsunori},
	month = sep,
	year = {2025},
	note = {arXiv:2509.14786 [cs]},
}

@misc{olmo_olmo_2025,
	title = {Olmo 3},
	url = {http://arxiv.org/abs/2512.13961},
	doi = {10.48550/arXiv.2512.13961},
	urldate = {2026-02-09},
	publisher = {arXiv},
	author = {Olmo, Team and Ettinger, Allyson and Bertsch, Amanda and Kuehl, Bailey and Graham, David and Heineman, David and Groeneveld, Dirk and Brahman, Faeze and Timbers, Finbarr and Ivison, Hamish and Morrison, Jacob and Poznanski, Jake and Lo, Kyle and Soldaini, Luca and Jordan, Matt and Chen, Mayee and Noukhovitch, Michael and Lambert, Nathan and Walsh, Pete and Dasigi, Pradeep and Berry, Robert and Malik, Saumya and Shah, Saurabh and Geng, Scott and Arora, Shane and Gupta, Shashank and Anderson, Taira and Xiao, Teng and Murray, Tyler and Romero, Tyler and Graf, Victoria and Asai, Akari and Bhagia, Akshita and Wettig, Alexander and Liu, Alisa and Rangapur, Aman and Anastasiades, Chloe and Huang, Costa and Schwenk, Dustin and Trivedi, Harsh and Magnusson, Ian and Lochner, Jaron and Liu, Jiacheng and Miranda, Lester James V. and Sap, Maarten and Morgan, Malia and Schmitz, Michael and Guerquin, Michal and Wilson, Michael and Huff, Regan and Bras, Ronan Le and Xin, Rui and Shao, Rulin and Skjonsberg, Sam and Shen, Shannon Zejiang and Li, Shuyue Stella and Wilde, Tucker and Pyatkin, Valentina and Merrill, Will and Chang, Yapei and Gu, Yuling and Zeng, Zhiyuan and Sabharwal, Ashish and Zettlemoyer, Luke and Koh, Pang Wei and Farhadi, Ali and Smith, Noah A. and Hajishirzi, Hannaneh},
	month = dec,
	year = {2025},
	note = {arXiv:2512.13961 [cs]},
}

@misc{muennighoff_scaling_2025,
	title = {Scaling {Data}-{Constrained} {Language} {Models}},
	url = {http://arxiv.org/abs/2305.16264},
	doi = {10.48550/arXiv.2305.16264},
	urldate = {2026-02-09},
	publisher = {arXiv},
	author = {Muennighoff, Niklas and Rush, Alexander M. and Barak, Boaz and Scao, Teven Le and Piktus, Aleksandra and Tazi, Nouamane and Pyysalo, Sampo and Wolf, Thomas and Raffel, Colin},
	month = jun,
	year = {2025},
	note = {arXiv:2305.16264 [cs]},
}

@misc{kaplan_scaling_2020,
	title = {Scaling {Laws} for {Neural} {Language} {Models}},
	url = {http://arxiv.org/abs/2001.08361},
	doi = {10.48550/arXiv.2001.08361},
	urldate = {2026-02-09},
	publisher = {arXiv},
	author = {Kaplan, Jared and McCandlish, Sam and Henighan, Tom and Brown, Tom B. and Chess, Benjamin and Child, Rewon and Gray, Scott and Radford, Alec and Wu, Jeffrey and Amodei, Dario},
	month = jan,
	year = {2020},
	note = {arXiv:2001.08361 [cs]},
}

@misc{hoffmann_training_2022,
	title = {Training {Compute}-{Optimal} {Large} {Language} {Models}},
	url = {http://arxiv.org/abs/2203.15556},
	doi = {10.48550/arXiv.2203.15556},
	urldate = {2026-02-09},
	publisher = {arXiv},
	author = {Hoffmann, Jordan and Borgeaud, Sebastian and Mensch, Arthur and Buchatskaya, Elena and Cai, Trevor and Rutherford, Eliza and Casas, Diego de Las and Hendricks, Lisa Anne and Welbl, Johannes and Clark, Aidan and Hennigan, Tom and Noland, Eric and Millican, Katie and Driessche, George van den and Damoc, Bogdan and Guy, Aurelia and Osindero, Simon and Simonyan, Karen and Elsen, Erich and Rae, Jack W. and Vinyals, Oriol and Sifre, Laurent},
	month = mar,
	year = {2022},
	note = {arXiv:2203.15556 [cs]},
}

@misc{datologyai_beyondweb_2025,
	title = {{BeyondWeb}: {Lessons} from {Scaling} {Synthetic} {Data} for {Trillion}-scale {Pretraining}},
	shorttitle = {{BeyondWeb}},
	url = {http://arxiv.org/abs/2508.10975},
	doi = {10.48550/arXiv.2508.10975},
	urldate = {2026-02-09},
	publisher = {arXiv},
	author = {DatologyAI and Maini, Pratyush and Dorna, Vineeth and Doshi, Parth and Carranza, Aldo and Pan, Fan and Urbanek, Jack and Burstein, Paul and Fang, Alex and Deng, Alvin and Abbas, Amro and Larsen, Brett and Blakeney, Cody and Bannur, Charvi and Baek, Christina and Teh, Darren and Schwab, David and Mongstad, Haakon and Yin, Haoli and Wills, Josh and Mentzer, Kaleigh and Merrick, Luke and Monti, Ricardo and Adiga, Rishabh and Joshi, Siddharth and Das, Spandan and Wang, Zhengping and Gaza, Bogdan and Morcos, Ari and Leavitt, Matthew},
	month = aug,
	year = {2025},
	note = {arXiv:2508.10975 [cs]},
}

@misc{kang_demystifying_2025,
	title = {Demystifying {Synthetic} {Data} in {LLM} {Pre}-training: {A} {Systematic} {Study} of {Scaling} {Laws}, {Benefits}, and {Pitfalls}},
	shorttitle = {Demystifying {Synthetic} {Data} in {LLM} {Pre}-training},
	url = {http://arxiv.org/abs/2510.01631},
	doi = {10.48550/arXiv.2510.01631},
	urldate = {2026-02-09},
	publisher = {arXiv},
	author = {Kang, Feiyang and Ardalani, Newsha and Kuchnik, Michael and Emad, Youssef and Elhoushi, Mostafa and Sengupta, Shubhabrata and Li, Shang-Wen and Raghavendra, Ramya and Jia, Ruoxi and Wu, Carole-Jean},
	month = oct,
	year = {2025},
	note = {arXiv:2510.01631 [cs]},
}


\clearpage
\appendix

\section{Implementation Details and Ablations for the CD-Scaling Fit}
\label{app:fit}

This appendix supplies the implementation details, the outlier-trimming sweep, the functional-form ranking, and the per-term ablation of $R^{*}$ referenced from \Cref{sec:fit_results,sec:ablations}.  The main text reports the canonical $k = 15$ row and the chosen exp-sat $R^{*}(\D, \N)$ form; the tables below show every cell of both sweeps so the choices can be audited.

\subsection{Implementation details}
\label{app:impl}

\paragraph{Log-space objective.} We work in log-loss space using the numerically stable LSE-equivalent reformulation of \Cref{eqn:data_chin}~\citep{hoffmann_training_2022}:
\begin{align}
    \log \Loss(\N, \D, \Dp) \;=\; \mathrm{LSE}\,\!\bigl(e,\; a - \alpha \log \N,\; b - \beta \log(\D + \eta \cdot \Dp)\bigr),
    \label{eq:lse_form}
\end{align}
algebraically identical to \Cref{eqn:data_chin} with $(\E, \A, \B) = (e^{e}, e^{a}, e^{b})$ but numerically stable. The fit minimises Huber loss on log-residuals with $\delta = 0.1$, which gives quadratic weight to typical residuals while suppressing the influence of stragglers an order of magnitude larger.

\paragraph{Optimiser and initialisation.} L-BFGS with strong-Wolfe line search.  We initialize from a log-spaced grid over the parameters, find the best in-sample seed, and report the optimum.  Across $30+$ random seeds the optimum is reproducible to four significant figures.

\paragraph{Joint fit.} The headline pipeline (\Cref{sec:fit_results}) fits all eleven parameters $(\E, \A, \B, \alpha, \beta)$ + $(\log K, \rho, \sigma)_{\text{rep}}$ + $(\log K, \rho, \sigma)_{\text{para}}$ jointly on the pooled 1-epoch + repetition + paraphrase data via a single Huber LSE with iterative residual trimming. The initialization grid brackets both signs of $\rho_{\text{para}}$ and $\sigma_{\text{para}}$ so the optimizer can land on either sign. Across every grid we tried, the optimum lands in the same negative-$(\rho_{\text{para}}, \sigma_{\text{para}})$ basin.

\paragraph{Iterative residual trimming.} A handful of noisy small-scale points can dominate the fit, so we drop them using an iterative variant of the residual-based trim of \citet{besiroglu2024chinchilla}: fit, drop the single worst residual, refit, drop the next worst, and so on for $k$ steps.  Applied to the untrimmed fit, the standard Hampel outlier rule \citep{hampel1974influence} identifies 14 points as outliers.  We round to $k = 15$ for consistency with our sweep grid (\Cref{tab:k_sweep}), which trims ${\sim}4\%$ of the data.  The trimmed points are all small-scale $1$-epoch and repetition runs at 14M and 30M, the same class of outliers identified by \citet{besiroglu2024chinchilla} for Chinchilla itself.


\subsection{Outlier-trimming sweep}
\label{app:k_sweep}


\begin{table}[h]
    \centering
       \caption{
    \textbf{Outlier-trimming sweep for the joint fit.} Iterative residual trimming on the pooled ($n = 356$) fit residuals.  Per-source RMSEs are on kept points.  The canonical $k = 15$ row (bold) is justified in \Cref{app:impl}.
    }
    \label{tab:k_sweep}
    \small
    \begin{tabular}{cccccccccc}
    \toprule
    $k$ & $n$ kept & $\beta$ & $\log K_{\text{rep}}$ & $\sigma_{\text{rep}}$ & $\log K_{\text{para}}$ & $\sigma_{\text{para}}$ & 1-ep RMSE & rep RMSE & para RMSE \\
    \midrule
      0 & 356 & 0.298 & 15.92 & $-0.63$ & 31.57 & $-1.34$ & 0.076 & 0.046 & 0.027 \\
      5 & 351 & 0.371 & 11.85 & $-0.44$ & 31.29 & $-1.32$ & 0.062 & 0.042 & 0.024 \\
     10 & 346 & 0.420 & 11.15 & $-0.41$ & 31.08 & $-1.31$ & 0.052 & 0.038 & 0.023 \\
     \textbf{15} & \textbf{341} & \textbf{0.437} & \textbf{10.96} & $\boldsymbol{-0.41}$ & \textbf{31.02} & $\boldsymbol{-1.33}$ & \textbf{0.043} & \textbf{0.035} & \textbf{0.024} \\
     20 & 336 & 0.456 & 10.82 & $-0.42$ & 31.02 & $-1.34$ & 0.039 & 0.031 & 0.024 \\
     25 & 331 & 0.472 & 10.71 & $-0.42$ & 31.03 & $-1.35$ & 0.035 & 0.029 & 0.022 \\
     30 & 326 & 0.482 & 10.64 & $-0.42$ & 31.03 & $-1.36$ & 0.030 & 0.027 & 0.020 \\
    \bottomrule
    \end{tabular}
    \vspace{.5em}
\end{table}

\subsection{Functional-form ranking for $\eta$}
\label{app:eta_forms}

We compared nine candidate forms for $\eta$ on the multi-epoch corpus, holding the backbone $(\B, \beta)$ fixed at the Stage-1 anchors and scoring each form by leave-one-out (LOO) RMSE on $\log \Loss$ (\Cref{tab:form_ranking}).  The three forms with a full $R^{*}(\D, \N)$ via \Cref{eq:rstar_anchor} (sat, exp-sat, tanh) dominate the rest at every parameter count.  We adopt exp-sat $R^{*}(\D, \N)$ (close 2nd in LOO behind tanh, $\Delta = 0.001$) because it is the canonical data-repetition form~\citep{muennighoff_scaling_2025}, has $\eta(0) = 1$ exactly, and approaches the $R^{*}$ asymptote monotonically from below.

\begin{table}[h]
    \centering
    \small
    \caption{
    \textbf{Functional-form ranking for $\eta$.}  LOO RMSE on $\log \Loss$ over $108$ multi-epoch points (kept after pooled residual trimming), with the backbone $(\B, \beta)$ frozen at the Stage-1 anchors.  The bottom three rows use the full $R^{*}(\D, \N)$ ansatz from \Cref{eq:rstar_anchor} ($\log R^{*} = \log K + \rho \log(\D/\N) + \sigma \log \N$).  Constant-$R^{*}$ (single-parameter Muennighoff baseline, $R^{*} = K$) is rejected by roughly $40\%$ relative LOO at the same parameter count, confirming that the $(\D, \N)$-dependence of $R^{*}$ is load-bearing.  We adopt exp-sat $R^{*}(\D, \N)$ (row in bold) as our functional form.
    }
    \label{tab:form_ranking}
    \begin{tabular}{lccc}
    \toprule
    Form & shape of $\eta \cdot r$ & $n_{\text{par}}$ & LOO RMSE \\
    \midrule
    constant $\eta$                                        & $c \, r$                                                      & 1 & 0.035 \\
    power in $r$                                           & $c \, r \cdot r^{-\gamma}$                                    & 2 & 0.032 \\
    sat in $r$                                             & $c \, r / (1 + b \, r)$                                       & 2 & 0.029 \\
    exp-decay, $R(\D/\N)$                                  & $\eta_{0} \, r \cdot e^{-r/R}$                                & 3 & 0.026 \\
    sat $\times (\D/\N)$, $b(\N)$                          & $c(\D/\N)^{-\gamma} r / (1 + b_{0}(\N/N_{0})^{\kappa} r)$      & 4 & 0.029 \\
    exp-sat, $R^{*}(\D/\N)$                                & $R^{*}(1 - e^{-r/R^{*}})$                                     & 2 & 0.026 \\
    sat, $R^{*}(\D, \N)$                                   & $R^{*} \, r / (R^{*} + r)$                                    & 3 & 0.021 \\
    \textbf{exp-sat, $R^{*}(\D, \N)$ (ours)}               & $R^{*}(1 - e^{-r/R^{*}})$                                     & \textbf{3} & \textbf{0.020} \\
    tanh, $R^{*}(\D, \N)$                                  & $R^{*} \tanh(r/R^{*})$                                        & 3 & 0.019 \\
    \bottomrule
    \end{tabular}
    \vspace{.2em}
\end{table}

\subsection{Per-term ablation of \Cref{eq:rstar_anchor}}
\label{app:term_ablation}

Holding the exp-sat shape of $\eta$ fixed, we ablate the terms of \Cref{eq:rstar_anchor} to test the necessity of each (\Cref{tab:term_ablation}).  Each exponent is individually necessary, and \Cref{fig:eta_analysis}(a, c) shows why: the $\D/\N$- and $\N$-dependence of $\eta$ are not absorbable into one another.

\begin{table}[h]
    \centering
    \small
    \caption{
    \textbf{Per-term ablation of $R^{*}$.} Starting from the original parametric form, we remove terms from $\log R^{*}$ and report the resulting LOO RMSE.  Removing either exponent individually degrades the fit, removing both recovers the constant-$R^{*}$~\citep{muennighoff_scaling_2025} baseline.
    }
    \label{tab:term_ablation}
    \begin{tabular}{l l c}
    \toprule
    Ablation & $\log R^{*}$ & LOO RMSE \\
    \midrule
    original                                                 & $\log K + \rho \log(\D/\N) + \sigma \log \N$ & \textbf{0.020} \\
    $\hookrightarrow$ removing $\N^{\sigma}$                 & $\log K + \rho \log(\D/\N)$                  & 0.026 \\
    $\hookrightarrow$ removing $(\D/\N)^{\rho}$              & $\log K + \sigma \log \N$                    & 0.025 \\
    $\hookrightarrow$ removing both   & $\log K$                                     & 0.028 \\
    \bottomrule
    \end{tabular}
    \vspace{.5em}
\end{table}

\subsection{Cross-scale validation across sizes}
\label{app:xval_sweep}

We now sweep the fit-set cutoff $N_{\max} \in \{30, 60, 100, 190\}$M and predict all held-out points at $\N > N_{\max}$ (\Cref{tab:xval_sweep}).

Held-out RMSE is stable across cutoffs between $0.05$ and $0.08$, and comparable to the in-sample RMSE at each cutoff.  Once $N_{\max} \geq 60$M, the recovered $\beta$ is within $0.03$ of the full-data value.  The CD-scaling law therefore extrapolates robustly across an order of magnitude in model size.

\begin{table}[h]
    \centering
    \small
    \caption{
    \textbf{Cross-scale validation across fit-set sizes.}  We refit on $\N \leq N_{\max}$ and predict all held-out $\N > N_{\max}$ points.  RMSE is reported on $\log \Loss$.
    }
    \label{tab:xval_sweep}
    \begin{tabular}{ccccccc}
    \toprule
    $N_{\max}$ & Fit sizes & $n_{\text{fit}}$ & $n_{\text{held}}$ & $\beta$ & in-sample RMSE & held-out RMSE \\
    \midrule
    $30$M   & 14M, 30M                    & 103 & 204 & 0.34 & 0.048 & 0.079 \\
    $60$M   & 14M, 30M, 60M               & 173 & 134 & 0.35 & 0.047 & $\mathbf{0.049}$ \\
    $100$M  & 14M, 30M, 60M, 100M         & 180 & 127 & 0.35 & 0.046 & $\mathbf{0.050}$ \\
    $190$M  & 14M, 30M, 60M, 100M, 190M   & 235 &  72 & 0.49 & 0.054 & $\mathbf{0.057}$ \\
    \bottomrule
    \end{tabular}
    \vspace{.5em}
\end{table}

\section{Additional Experiment Details and Results}

\subsection{Cross-scale validation across sizes}
\label{app:xval_sweep}

We now sweep the fit-set cutoff $N_{\max} \in \{30, 60, 100, 190\}$M and predict all held-out points at $\N > N_{\max}$ (\Cref{tab:xval_sweep}).

Held-out RMSE is stable across cutoffs between $0.05$ and $0.08$, and comparable to the in-sample RMSE at each cutoff.  Once $N_{\max} \geq 60$M, the recovered $\beta$ is within $0.03$ of the full-data value.  The CD-scaling law therefore extrapolates robustly across an order of magnitude in model size.

\begin{table}[h]
    \centering
    \small
    \caption{
    \textbf{Cross-scale validation across fit-set sizes.}  We refit on $\N \leq N_{\max}$ and predict all held-out $\N > N_{\max}$ points.  RMSE is reported on $\log \Loss$.
    }
    \label{tab:xval_sweep}
    \begin{tabular}{ccccccc}
    \toprule
    $N_{\max}$ & Fit sizes & $n_{\text{fit}}$ & $n_{\text{held}}$ & $\beta$ & in-sample RMSE & held-out RMSE \\
    \midrule
    $30$M   & 14M, 30M                    & 103 & 204 & 0.34 & 0.048 & 0.079 \\
    $60$M   & 14M, 30M, 60M               & 173 & 134 & 0.35 & 0.047 & $\mathbf{0.049}$ \\
    $100$M  & 14M, 30M, 60M, 100M         & 180 & 127 & 0.35 & 0.046 & $\mathbf{0.050}$ \\
    $190$M  & 14M, 30M, 60M, 100M, 190M   & 235 &  72 & 0.49 & 0.054 & $\mathbf{0.057}$ \\
    \bottomrule
    \end{tabular}
    \vspace{.5em}
\end{table}

\subsection{Loss to Downstream Breakdown}
\begin{figure}[H]
    \centering
    \includegraphics[width=1.0\linewidth]{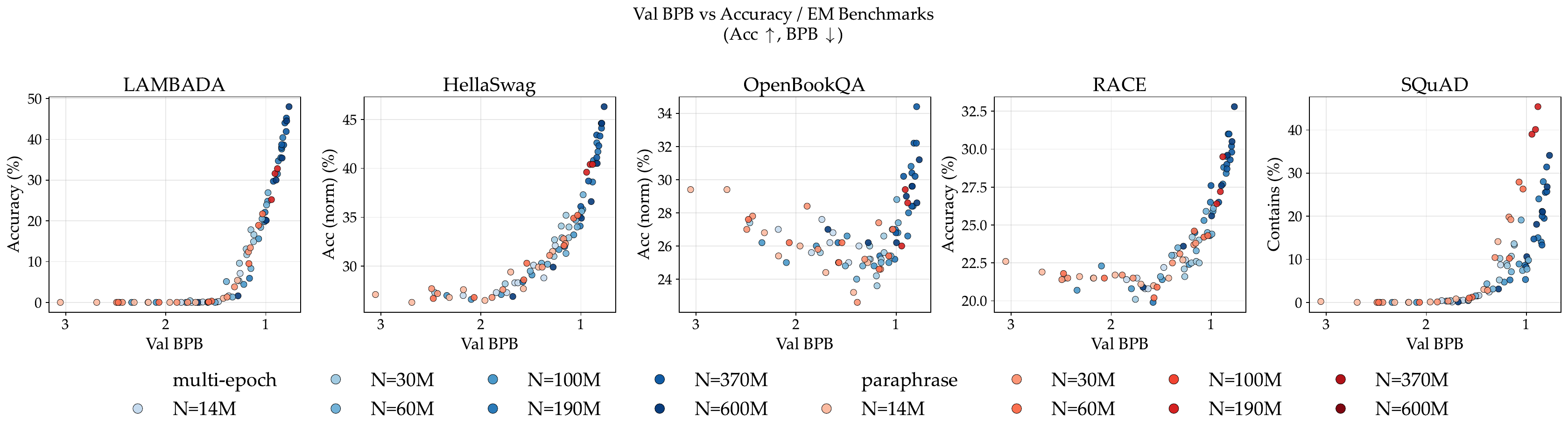}
    \caption{
    \textbf{Per-task breakdown of accuracy-based benchmarks versus validation BPB.} Each subplot shows one task against validation BPB, decomposing the accuracy side of \Cref{fig:loss_benchmark_main}. Cool hues mark multi-epoch runs and warm hues mark paraphrase runs, with color denoting model size $\N$ (14M to 600M). On every task, runs of all strategies and scales fall onto a single monotonic curve in validation BPB.
    }
    \label{fig:val_loss_vs_benchmarks_acc}
\end{figure}

\begin{figure}[H]
    \centering
    \includegraphics[width=1.0\linewidth]{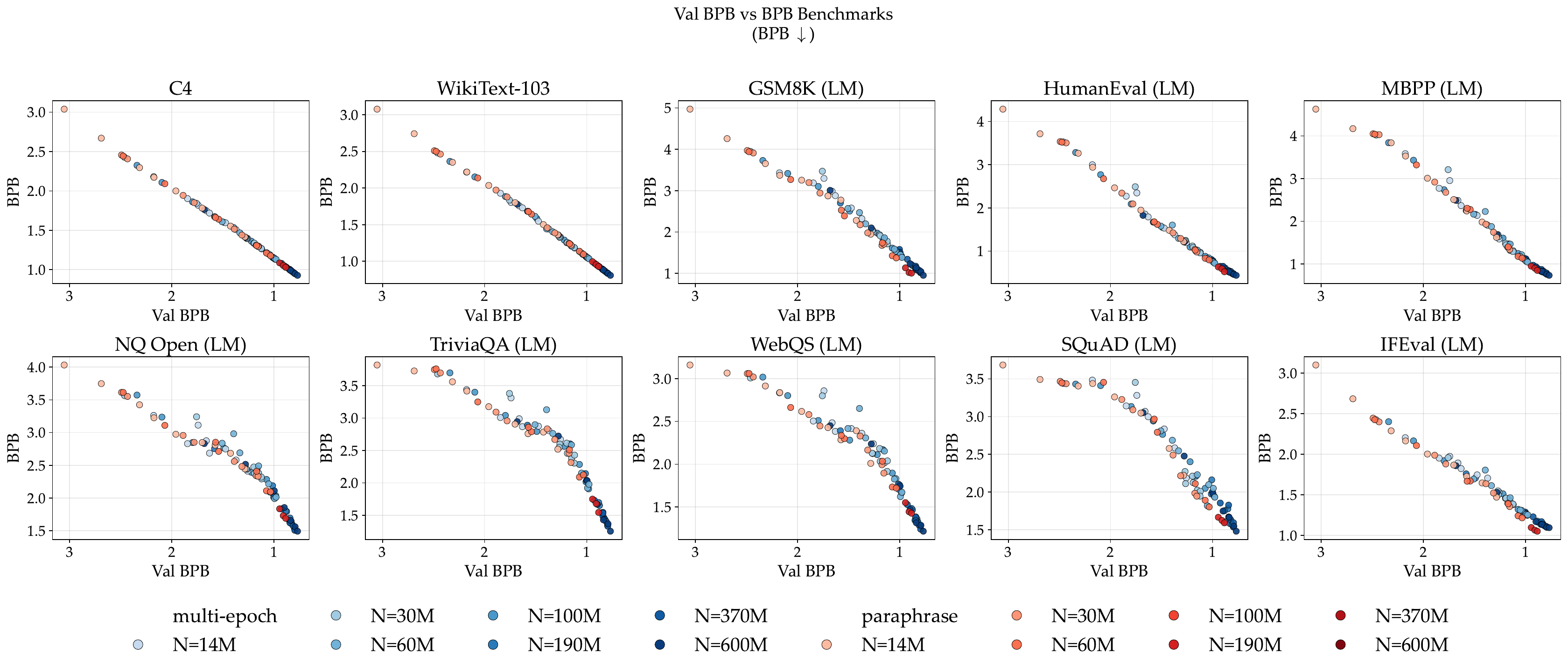}
    \caption{
    \textbf{Per-task breakdown of BPB-based benchmarks versus validation BPB.} Each subplot shows one task (recast as language modeling and scored in BPB) against validation BPB, decomposing the BPB side of \Cref{fig:loss_benchmark_main}. Cool hues mark multi-epoch runs and warm hues mark paraphrase runs, with color denoting model size $\N$ (14M to 600M). On every task, runs of all strategies and scales fall onto a single monotonic curve in validation BPB.
    }
    \label{fig:val_loss_vs_benchmarks_bpb}
\end{figure}

In \Cref{fig:val_loss_vs_benchmarks_acc}, and \Cref{fig:val_loss_vs_benchmarks_bpb}, we report loss-to-downstream for each benchmark. 

\label{app:loss_benchmark_breakdown}

\subsection{Optimal Scaling Path for 370M model}
\begin{figure}[h]
    \centering
    \includegraphics[width=1.0\linewidth]{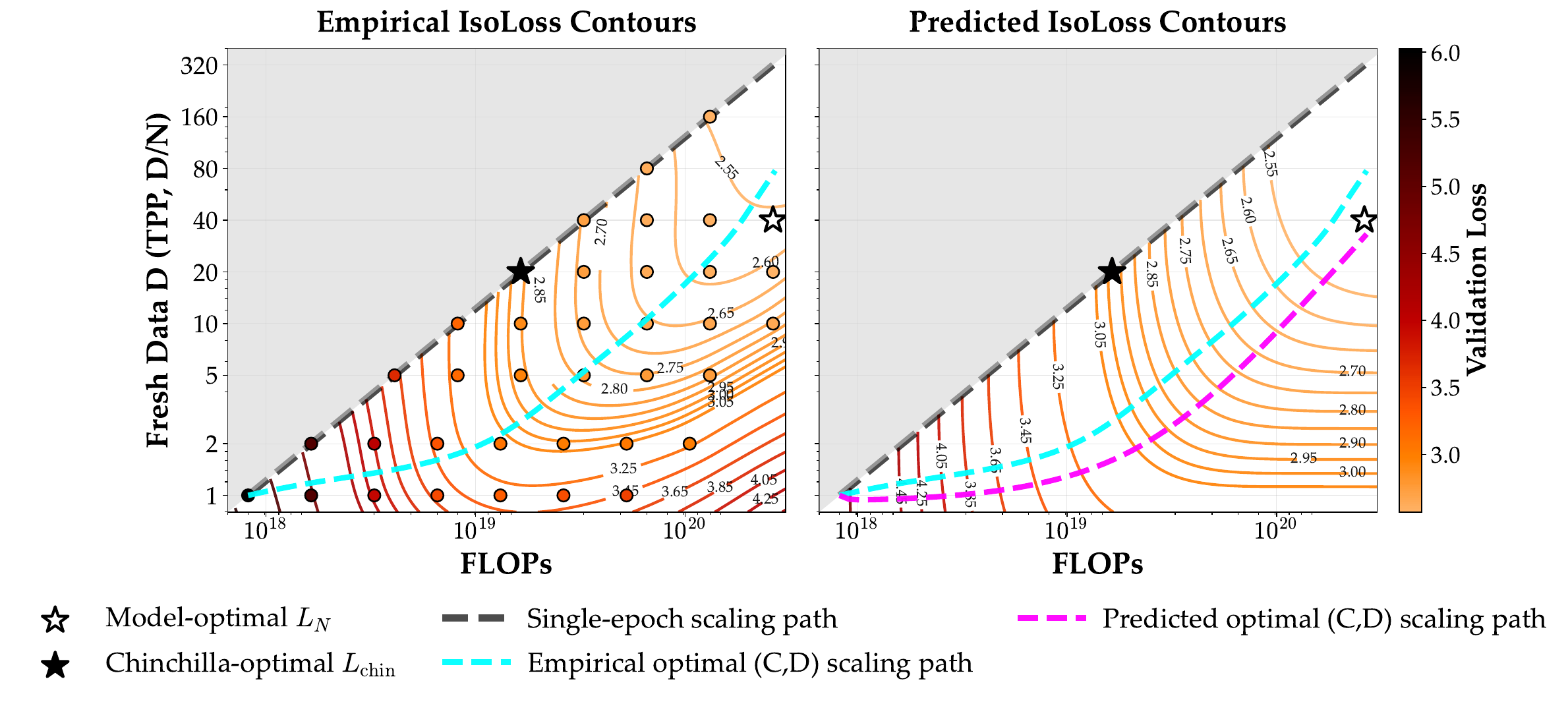}
    \caption{
    \textbf{Optimal compute-data allocation on (\C, \D) loss surfaces for $\N=370M$.} Replicating results for \Cref{fig:multi_epoch_scaling}. 
    }
    \label{fig:loss_surface_370m}
\end{figure}

In \Cref{fig:loss_surface_370m}, we visualize the optimal scaling path for 370M model and show that the scaling law path predicted by CD-scaling laws also follow closely with the empirical optimal path. 

\subsection{Training Run Details}

\paragraph{Hyperparameter Sweep.}
We visualize the hyperparamter sweep grid and validation loss for 30M, 370M models. Lower epoch requires smaller weight-decay and as we increase data repetition, higher weight decay is necessary. For larger epochs, 30M model requires WD as high as 1.6, which is consistent with observations made in \citep{kim_pre-training_2025}. 

\paragraph{Other Hyperparameters.}
While we sweep learning rate and weight decay for our training runs, we list the rest of hyperparameter in \Cref{tab:hyperparams_pretrain}.

\begin{figure}[h]
    \centering
    \includegraphics[width=1.0\linewidth]{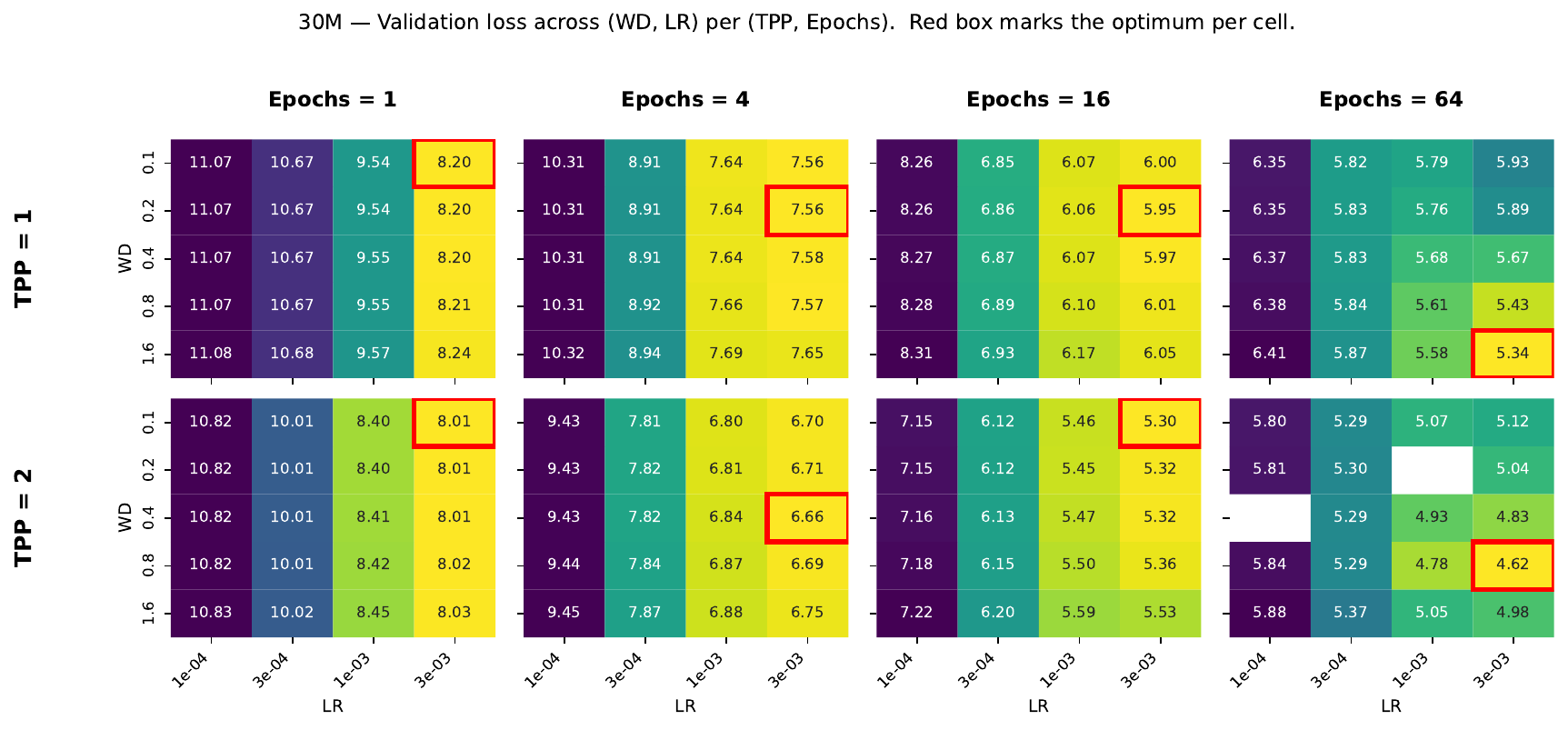}
    \includegraphics[width=1.0\linewidth]{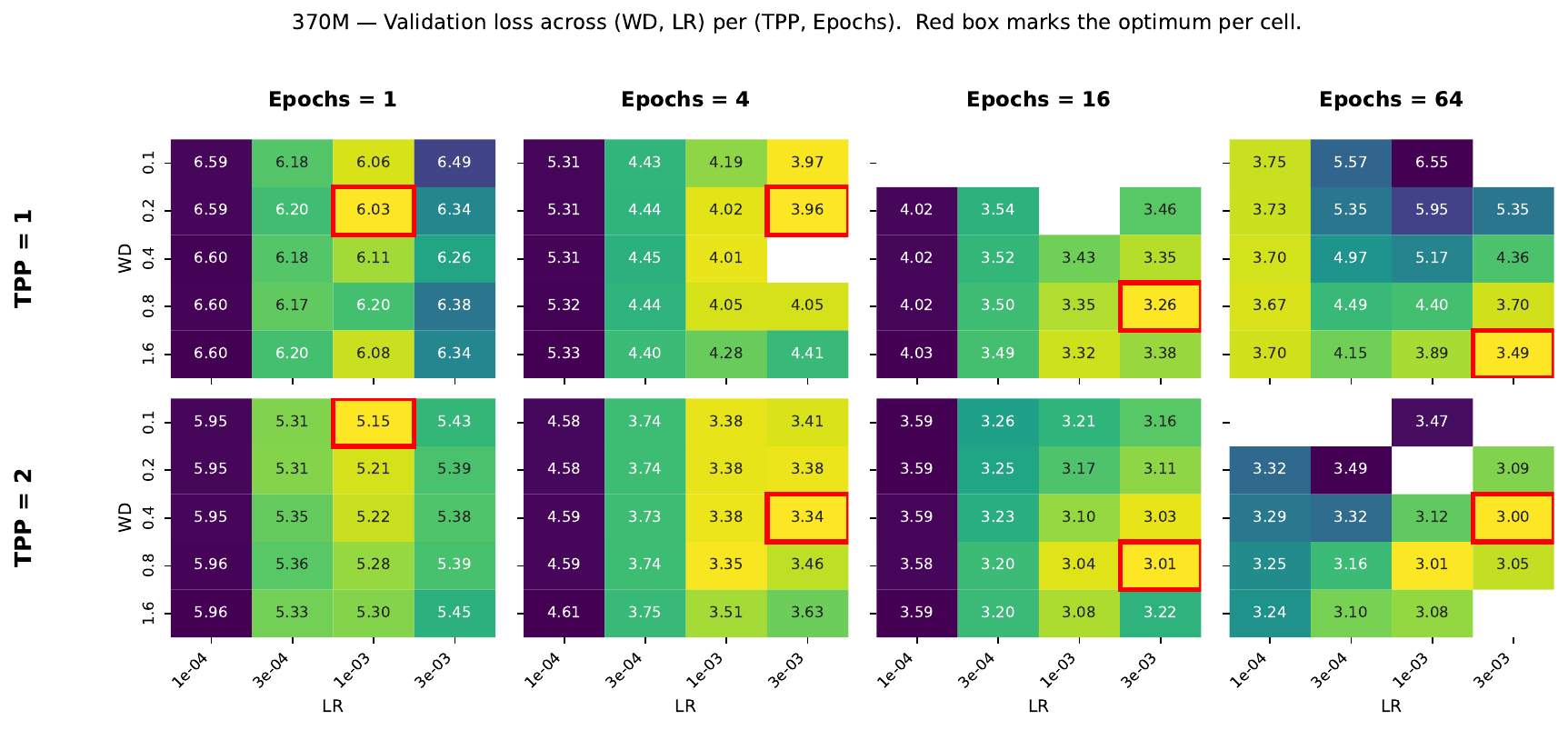}
    \caption{
    \textbf{Hyperparameter grid example for 30M and 370M model.} More data repetition requires much higher weight decay especially for small models. 
    }
    \label{fig:hparam}
\end{figure}

\begin{table}[htbp]
\centering
\caption{Pretraining hyperparameters (OLMo2-1B, 50B tokens).}
\label{tab:hyperparams_pretrain}
\begin{tabular}{lll}
\toprule
\textbf{Category} & \textbf{Hyperparameter} & \textbf{Value} \\
\midrule
\multirow{3}{*}{Data}
  & Global batch size (sequences) & 512 \\
  & Gradient accumulation steps  & 64 \\
\cmidrule{1-3}
\multirow{5}{*}{Optimization}
  & Learning rate                & tuned \\
  & Optimizer                    & AdamW \\
  & $(\beta_1, \beta_2)$         & $(0.9,\ 0.95)$ \\
  & Weight decay                 & tuned \\
  & Gradient clip                & 1.0 \\
\cmidrule{1-3}
\multirow{4}{*}{LR Schedule}
  & Schedule                     & cosine with warmup \\
  & Warmup tokens                & 1B \\
  & Min LR ratio ($\alpha_f$)    & 0.1 \\
  & Units                        & tokens \\
\cmidrule{1-3}
\multirow{3}{*}{Regularization}
  & Precision                    & BF16 (AMP) \\
  & Softmax auxiliary loss       & \checkmark \\
  & Auxiliary loss multiplier    & $1 \times 10^{-5}$ \\
\bottomrule
\end{tabular}
\end{table}


\end{document}